\documentclass[sn-nature]{sn-jnl}

\usepackage{graphicx}%
\usepackage{multirow}%
\usepackage{amsmath,amssymb,amsfonts}%
\usepackage{amsthm}%
\usepackage{mathrsfs}%
\usepackage[title]{appendix}%
\usepackage{xcolor}%
\usepackage{textcomp}%
\usepackage{manyfoot}%
\usepackage{booktabs}%
\usepackage{algorithm}
\usepackage{algpseudocode}

\usepackage{float}
\usepackage{subcaption}
\usepackage{listings}%
\usepackage{multirow}

\usepackage{tikz}
\usetikzlibrary{positioning, shapes.geometric}
\usepackage{array}
\usepackage{longtable}
\usepackage{amsmath}

\theoremstyle{thmstyleone}%
\theoremstyle{thmstyletwo}%

\theoremstyle{thmstylethree}%
\newtheorem{definition}{Definition}[section]%
\newtheorem{problem}{Problem}
\usepackage{xcolor}
\begin{document}

\title[]{Heterogeneous Graph Auto-Encoder for Credit Card Fraud Detection}

\author[1]{\fnm{Moirangthem Tiken} \sur{Singh}}\email{tiken.m@dibru.ac.in}

 \author[2]{\fnm{Rabinder Kumar} \sur{Prasad}}\email{rkp@dibru.ac.in}

\author[3]{\fnm{Gurumayum Robert} \sur{Michael}}\email{robertmichael@dibru.ac.in}

 \author[4]{\fnm{N K} \sur{Kaphungkui}}\email{pipizs.kaps@gmail.com}

 \author[5]{\fnm{N.Hemarjit} \sur{Singh}}\email{nhsingh@dibru.ac.in}

\affil[1,2]{\orgdiv{Department of Computer Science and Engineering}, \orgname{Dibrugarh University Institute of Engineering and Technology}, \orgaddress{\street{Dibrugarh University}, \city{Dibrugarh}, \postcode{786004}, \state{Assam}, \country{India}}}
\affil[3,4,5]{\orgdiv{Department of Electronic and Communication Engineering}, \orgname{Dibrugarh University Institute of Engineering and Technology}, \orgaddress{\street{Dibrugarh University}, \city{Dibrugarh}, \postcode{786004}, \state{Assam}, \country{India}}}


\abstract{
The digital revolution has significantly impacted financial transactions, leading to a notable increase in credit card usage. However, this convenience comes with a trade-off: a substantial rise in fraudulent activities. Traditional machine learning methods for fraud detection often struggle to capture the inherent interconnectedness within financial data. This paper proposes a novel approach for credit card fraud detection that leverages Graph Neural Networks (GNNs) with attention mechanisms applied to heterogeneous graph representations of financial data. Unlike homogeneous graphs, heterogeneous graphs capture intricate relationships between various entities in the financial ecosystem, such as cardholders, merchants, and transactions, providing a richer and more comprehensive data representation for fraud analysis. To address the inherent class imbalance in fraud data, where genuine transactions significantly outnumber fraudulent ones, the proposed approach integrates an autoencoder. This autoencoder, trained on genuine transactions, learns a latent representation and flags deviations during reconstruction as potential fraud. This research investigates two key questions: (1) How effectively can a GNN with an attention mechanism detect and prevent credit card fraud when applied to a heterogeneous graph? (2) How does the efficacy of the autoencoder with attention approach compare to traditional methods? The results are promising, demonstrating that the proposed model outperforms benchmark algorithms such as Graph Sage and FI-GRL, achieving a superior AUC-PR of 0.89 and an F1-score of 0.81. This research significantly advances fraud detection systems and the overall security of financial transactions by leveraging GNNs with attention mechanisms and addressing class imbalance through an autoencoder.}





\maketitle

\section{Introduction}

Financial transactions, especially credit card usage, have experienced a surge due to the digital revolution. This has resulted in a vast amount of financial data, empowering companies to comprehend customer behavior and utilize data for decision-making. On the other hand, the convenience that comes with this has a downside - there is a noticeable rise in fraudulent activities. Traditional methods of fraud detection often struggle to keep pace with the evolving nature of these schemes. In order to tackle this challenge, the field of machine learning (ML) has surfaced as a potent tool that can effectively identify and prevent fraudulent transactions \cite{Financial2022}. By leveraging ML algorithms, it becomes possible to analyze massive amounts of financial data, identify recurring patterns, and pinpoint potential fraud through anomaly detection. They enable financial institutions to automate the fraud detection process, facilitating real-time monitoring of transactions and activities. To detect fraud effectively, many professionals rely on techniques such as decision trees, random forests, and support vector machines \cite{hussain2021fraud,9540093}.

The conventional approaches to detecting fraud often face difficulties in capturing the intrinsic interrelationships that exist within financial data. Transactions typically involve multiple parties, including cardholders, merchants, banks, and various other entities. The representation of financial transactions as a graph enables us to take advantage of the connections among them, thereby enhancing the effectiveness of fraud detection measures. Despite their widespread use, it is important to acknowledge that traditional methods may face difficulties in accurately differentiating between relevant and irrelevant relationships within the graph, thus impacting their ability to effectively detect fraudulent activity.

Graph Neural Networks (GNNs) excel at processing graph data and utilizing attention mechanisms to focus on the most relevant entities and relationships within the network structure \cite{zhou2020graph}. This makes them well-suited for tasks like fraud detection, where identifying the most critical factors contributing to a transaction's legitimacy is crucial. By applying attention, the GNN can prioritize information from neighboring nodes (e.g., cardholder's spending habits, merchant's location) that are most relevant to understanding the transaction's nature. This refined focus on critical relationships improves the model's ability to distinguish between normal transactions and those exhibiting suspicious patterns, potentially indicative of fraud.

 In graph induction learning techniques, two types of graph representations of data are used: homogeneous graph \cite{Bo2023} and heterogeneous graph \cite{Shi2022}. Financial fraud data, especially involving credit cards, is inherently heterogeneous. It encompasses diverse entities like cardholders, merchants, and transactions, each with distinct attributes and relationships. Homogeneous graphs, which represent entities of the same type, may not fully capture this complexity. In contrast, heterogeneous graphs offer a more comprehensive representation, effectively capturing the multifaceted nature of financial transactions and the intricate relationships between entities within the financial ecosystem.

\begin{figure}[h!]
    \centering
    \includegraphics[width=0.6\textwidth]{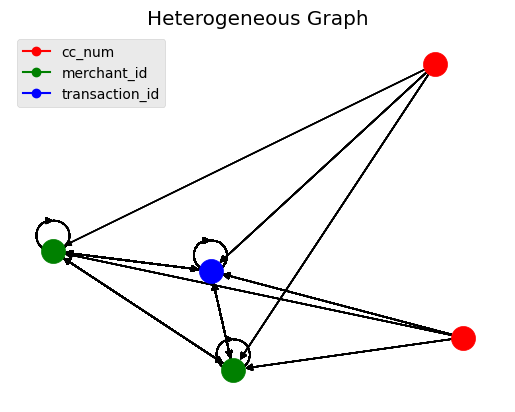}
    \caption{Relationships between different nodes.}
    \label{fig:result01}
\end{figure}

 For instance, a heterogeneous graph might include nodes representing credit card numbers (\texttt{cc\_num}), merchants information (\texttt{merchant\_id}), and transaction numbers (\texttt{transaction\_id}), with edges connecting them based on the specific relationship (e.g., a transaction between a cardholder and a merchant). This allows us to analyze the network structure and identify suspicious patterns that might be missed by simpler models. For instance, in Figure~\ref{fig:result01}, the relationships between different data points are illustrated. These relationships are often overlooked by homogeneous graph learning algorithms and their variants.

The varying characteristics of nodes and edges in heterogeneous graph data make it difficult to apply GNNs directly, thereby necessitating a more sophisticated approach for information aggregation than what is typically used for homogeneous graphs. In addition, the effectiveness of supervised learning is often hindered by class imbalance in fraud data. This imbalance is characterized by a significantly smaller number of fraudulent transactions compared to genuine transactions. As a result, traditional supervised learning models struggle to learn effectively from such imbalanced data \cite{ijfs11030110}.

This work suggests a new approach that effectively handles heterogeneous graph data by leveraging advanced GNN techniques for aggregating information from diverse node and edge types. These techniques ensure that the varying attributes and relationships within the graph are adequately captured and utilized in the analysis process.

Furthermore, to tackle the issue of class imbalance, common techniques such as oversampling and undersampling \cite{brownlee2021combine} are used. Balancing class distribution can be achieved through oversampling, which generates more instances of the minority class (fraud transactions), or through undersampling, which reduces instances of the majority class (genuine transactions). Nonetheless, these approaches may be complicated and possess their own limitations.

To overcome these challenges, this approach integrates an autoencoder (AE) with a decoder that is trained on genuine transactions. By learning a latent representation, the AE can accurately reconstruct these transactions. The ability to detect fraudulent activities in complex heterogeneous graph data is enhanced by flagging deviations from the learned distribution during reconstruction, thereby addressing class imbalance.

Considering all scenarios discussed, this work aims to answer the following research questions (RQs):
\begin{itemize}
	\item {\textbf{RQ1: Effectiveness of GNNs with Attention for Fraud Detection:}}How effectively can GNNs utilizing an attention mechanism detect and prevent credit card fraud when applied to a heterogeneous graph representation that captures the complex interrelationships within the financial ecosystem?
	\item {\textbf{RQ2: Comparison of Autoencoder with Attention vs. Traditional Methods:}} How does the proposed autoencoder-based fraud detection approach, which leverages GNNs with attention and is trained on a non-fraudulent transaction graph dataset, compare to traditional methods in terms of accuracy, efficiency, and scalability, especially considering significant class imbalance?
\end{itemize}

The methodology consists of several steps, one of which is the processing of a tabular dataset of financial transactions. This dataset is then transformed into a heterogeneous graph. As a result, the graph is subjected to analysis using autoencoders (AE) and graph neural networks (GNNs), which enables the identification of anomalies that can be linked to fraudulent activity. By focusing on the class imbalance problem, the proposed approach effectively tackles the challenge of fraud detection tasks. The results of this work have significant implications for businesses and financial institutions, empowering them to gain valuable insights into customer behavior and enhance their ability to identify and prevent fraudulent transactions. Ultimately, this work contributes to the advancement of fraud detection systems and the overall security of financial transactions in the digital era.

This paper provides a comprehensive discussion of the relevant literature in Section \ref{lit}. The problem statement is outlined in Section \ref{ps}, aiming to address a specific problem. The methodology employed in this research is elucidated in Section \ref{methodology}. The results obtained from this methodology are analyzed and presented in Section \ref{imp-res}. Finally, Section \ref{conclu} concludes the paper by summarizing the key findings and implications.

\section{Literature Review}\label{lit}

In this section, we introduce a range of notable works that cover various topics such as probabilistic graphical models, machine learning algorithms (including deep learning models), and advanced graph neural networks and their various variants. Table \ref{table:lit} provides a summary of the articles related to the proposed model.

 Papers such as \cite{4358713} and \cite{ROBINSON2018235} aim to address the problem of fraud detection in credit card transactions by modeling these transactions using a Hidden Markov Model (HMM), a probabilistic graphical model. The primary difference between them lies in their approach: in the first paper, a card-centric HMM is employed to detect abnormalities in transactions, while the latter paper opts for a merchant-centric HMM model. Both methods have the capability to identify fraud in real-time for merchants, operating in conjunction with modern transaction processing systems that handle card transactions.

 Additionally, \cite{LUCAS2020393} models credit card transaction sequences using the HMM approach, considering three distinct perspectives:

(i) Determining whether fraud is present or absent in the sequence.

(ii) Crafting sequences by fixing either the cardholder or the payment terminal.

(iii) Constructing sequences based on the spent amounts or the elapsed time between consecutive transactions. The combination of these three binary perspectives results in eight distinct sets of sequences derived from the training dataset of transactions. Each of these sequences is then represented using a Hidden Markov Model (HMM). Subsequently, each HMM assigns a likelihood to a transaction based on its sequence of preceding transactions. These likelihood values serve as additional features for the Random Forest classifier to detect fraud. In brief, this model provides a concept of sequential information flow during credit card transactions as part of a feature for a machine learning model.

The paper \cite{Itoo2021} explores the issue of credit card fraud detection and conducts a comparative analysis of three machine learning algorithms: logistic regression, Naïve Bayes, and K-nearest neighbor. To address the class imbalance, the authors utilize different proportions of the dataset and employ a random undersampling technique. They evaluate the algorithms based on various metrics. According to the results, the logistic regression-based model outperforms the prediction models derived from Naïve Bayes and K-nearest neighbor. The paper also suggests that applying undersampling techniques to the data before model development can lead to improved results. In addition, several machine learning algorithms, such as support vector machine (SVM) \cite{9640631}, random forest (RF) \cite{9640631,math9212683}, AdaBoost, and Majority Voting \cite{8292883}, as well as artificial neural network (ANN) \cite{RB202135,10.1007/978-3-030-89654-6_2}, are being explored as models for controlling fraudulent transactions in credit cards.

To enhance the performance of the above-mentioned models, \cite{Ileberi2022} defines a model in an ML-driven credit card fraud detection system that uses the genetic algorithm (GA) for feature selection. After identifying optimal features, this detection system utilizes a range of ML classifiers, including Decision Tree (DT), Random Forest (RF), Logistic Regression (LR), Artificial Neural Network (ANN), and Naive Bayes (NB).

While the aforementioned models perform well, a significant class imbalance exists in the credit card fraud dataset, with non-fraudulent transactions vastly outnumbering fraudulent ones. As a result, these models tend to prioritize high precision by predominantly predicting the majority class. To address this issue, several machine learning models (referenced as \cite{10081315}) employ one or a combination of oversampling and undersampling techniques (as mentioned in \cite{10.1145/2907070}). 

The study cited as \cite{doi:10.1080/0952813X.2021.1907795} conducts a comparative investigation of various approaches to address class imbalance. The findings indicate that a combination of oversampling and undersampling methods performs well when applied to ensemble classification models, including AdaBoost, XGBoost, and Random Forest. Deep learning algorithms such as Long Short-Term Memory (LSTM) and Gated Recurrent Unit (GRU), combined with a multilayer perceptron, are employed in the studies referenced as \cite{10081315} and \cite{9698195}. In \cite{9698195}, the authors use the Hybrid Synthetic Minority Oversampling Technique and Edited Nearest Neighbor (SMOTE-ENN) to balance the distribution of positive (fraud) and negative (non-fraud) instances in the dataset. However, the effectiveness of the SMOTE-ENN technique is crucial, as poor performance in resampling can significantly degrade the model's overall performance.

While oversampling and undersampling techniques can address class imbalance, they come with drawbacks like increased computational cost, potential for overfitting, and information loss (as discussed in \cite{brownlee2021combine, Yang2022}). Additionally, they can be sensitive to noise \cite{Zhang2023} and have limited effectiveness for highly imbalanced datasets \cite{ebiaredoh2022machine}. Therefore, \cite{ebiaredoh2022machine} propose an approach for Chronic Kidney Disease (CKD) prediction using imbalanced data. Their method leverages information gain-based feature selection and a cost-sensitive AdaBoost classifier. However, this approach focuses on spatial data and might not be suitable for graph data due to potential loss of structural information and inadequate feature representation during feature selection. So, such models will often struggle to capture the full picture of fraudulent activity. As noted in \cite{9204584}, many methods focus solely on spatial data points representing financial transactions, neglecting the valuable insights from temporal relationships. This limitation hinders the ability of these models to identify evolving fraud patterns. Furthermore, many existing models rely solely on labeled data for training, restricting their ability to leverage the vast amount of unlabeled data available in real-world credit card transactions \cite{sheng2023semi}.

To address these issues, an increasing number of researchers are exploring graph-based techniques for fraud detection, as discussed in \cite{9204584} and \cite{9724422}. In this approach, datasets are transformed into graphs, providing a better understanding of the relationships among financial transactions. Graph Neural Network (GNN) algorithms, as detailed in \cite{zhang2023expressive}, are applied to these graph datasets, allowing for efficient data aggregation from neighboring nodes and the extraction of node representations within the graph datasets. Among the popular GNN variants, GraphSAGE \cite{hamilton2018inductive} and GAT \cite{vel2018graph} stand out, utilizing sampling methods and attention mechanisms to gather neighbor information. These techniques have shown promising results in the field of fraud detection. Furthermore, the paper \cite {10.1145/3442381.3449989} introduces an algorithm designed to tackle the class imbalance problem in graph-based fraud detection. It employs an algorithm known as Pick and Choose Graph Neural Network (PC-GNN) to perform imbalanced supervised learning on graphs. The PC-GNN algorithm selects neighbor candidates for each node within the sub-graph using a neighborhood sampler. Ultimately, it aggregates information from the chosen neighbors and different relations to derive the final representation of a target node. The paper reports that PC-GNN surpasses state-of-the-art baselines in both benchmark and real-world graph-based fraud detection tasks.

However, inconsistency issues arise in the aggregation process of GNN models when applied to fraud detection tasks \cite{10.1145/3397271.3401253}. The aggregation mechanism relies on the assumption that neighbors share similar features and labels. When this assumption breaks down, the aggregation of neighborhood information becomes ineffective in learning node embeddings.

 To address these challenges, researchers in \cite{10.1145/3397271.3401253} and \cite{10.1145/3623401} have employed a multi-relational graph, known as a heterogeneous graph, for the classification of financial fraud. In \cite{10.1145/3397271.3401253}, context inconsistency, feature inconsistency, and relation inconsistency in GNN are introduced. To tackle these inconsistencies, the authors propose a new GNN framework called GraphConsis. GraphConsis addresses these issues by combining context embeddings with node features to handle context inconsistency, designing a consistency score to filter inconsistent neighbors and generate corresponding sampling probabilities to address feature inconsistency, and learning relation attention weights associated with the sampled nodes to tackle relation inconsistency.

In \cite{10.1145/3623401}, the authors propose semi-supervised methods that operate with heterogeneous graph datasets to address class imbalance issues in online credit loans. This paper utilizes a Graph-Oriented Snorkel approach to incorporate external expert knowledge, ultimately improving the performance of the learning algorithm when dealing with imbalanced datasets.

Another noteworthy work, \cite{10.1145/3269206.3272010}, introduces a heterogeneous graph-based approach for detecting malicious accounts in financial transactions. The authors present an algorithm called GEM, which adapts to learn discriminative embeddings for various node types. GEM employs an aggregator to capture node patterns within each type and utilizes an attention mechanism to enhance algorithm efficiency.

In \cite{10.14778/3494124.3494128}, the authors endeavor to design heterogeneous graph embeddings. Their approach incorporates heterogeneous mutual attention and heterogeneous message passing, incorporating key, value, and query vector operations (self-attention mechanism). This work features both a detector and an explainer, capable of predicting the validity of incoming transactions and providing insightful, understandable explanations generated from graphs to aid in subsequent business unit procedures.

The framework employed in \cite{VANBELLE2022116463} utilizes an algorithm for graph representation learning to create concise numerical vectors that capture the underlying network structure. The authors in this work assess the predictive capabilities of inductive graph representation learning with GraphSage and Fast Inductive Graph Representation Learning algorithms on credit card datasets characterized by significant data imbalance.

	\begin{longtable}{| m{3cm} | m{3.5cm} | m{3.5cm} | m{3.5cm} |}
		\hline
		\textbf{Model/Technique} & \textbf{Description} & \textbf{Strengths} & \textbf{Weaknesses} \\
		\hline
		\endfirsthead
		
		\multicolumn{4}{c}%
		{{\bfseries \tablename\ \thetable{} -- continued from previous page}} \\
		\hline
		\textbf{Model/Technique} & \textbf{Description} & \textbf{Strengths} & \textbf{Weaknesses} \\
		\hline
		\endhead
		
		\hline \multicolumn{4}{|r|}{{Continued on next page}} \\ \hline
		\endfoot
		
		\hline
		\endlastfoot
		
		Card-centric HMM \cite{4358713} & Focuses on cardholder transaction patterns. & Specific to individual cardholder behavior. & Limited by card-specific anomalies. \\
		\hline
		Merchant-centric HMM \cite{ROBINSON2018235} & Focuses on merchant transaction patterns. & Captures merchant-specific fraud patterns. & May miss cardholder-specific fraud patterns. \\
		\hline
		HMM + Random Forest \cite{LUCAS2020393} & Combines HMM for sequence modeling with Random Forest for classification. & Incorporates sequential information into machine learning. & Complex feature engineering required. \\
		\hline
		Logistic Regression \cite{Itoo2021} & Standard ML algorithm for binary classification. & Simplicity, interpretability, outperforms Naïve Bayes and KNN in study. & Limited capacity to handle complex patterns. \\
		\hline
		Naïve Bayes \cite{Itoo2021} & Probabilistic classifier based on Bayes' theorem. & Fast, easy to implement. & Assumes feature independence, less effective for complex data. \\
		\hline
		K-nearest Neighbor (KNN) \cite{Itoo2021} & Instance-based learning algorithm. & Simple, effective for small datasets. & Computationally expensive, less effective on large/imbalanced datasets. \\
		\hline
		Support Vector Machine (SVM) \cite{9640631} & Supervised learning model for classification. & Effective in high-dimensional spaces. & Memory-intensive, challenging with large datasets. \\
		\hline
		Random Forest (RF) \cite{9640631,math9212683} & Ensemble learning method using multiple decision trees. & High accuracy, robustness to overfitting. & May require extensive computational resources. \\
		\hline
		AdaBoost \cite{8292883} & Boosting algorithm combining weak classifiers. & Improves model performance by focusing on misclassified instances. & Sensitive to noisy data and outliers. \\
		\hline
		Artificial Neural Network (ANN) \cite{RB202135,10.1007/978-3-030-89654-6_2} & Deep learning models for complex pattern recognition. & Can model complex relationships, high accuracy. & Requires large datasets, computationally intensive. \\
		\hline
		Genetic Algorithm (GA) + ML \cite{Ileberi2022} & Uses GA for feature selection, combined with various ML classifiers. & Optimizes feature set for better model performance. & Computationally expensive, complex implementation. \\
		\hline
		Hybrid SMOTE-ENN \cite{10081315,9698195} & Combines Synthetic Minority Oversampling Technique and Edited Nearest Neighbor for class balancing. & Balances dataset effectively, enhances model performance. & Computationally intensive, potential sensitivity to noise. \\
		\hline
		Graph Neural Networks (GNN) \cite{zhang2023expressive} & Leverages graph structures for node representation and classification. & Captures relational data effectively, scalable with GraphSAGE and GAT. & Inconsistency issues in node aggregation, complexity in implementation. \\
		\hline
		PC-GNN \cite{10.1145/3442381.3449989} & Pick and Choose GNN for imbalanced learning on graphs. & Effective for imbalanced graph datasets, state-of-the-art performance. & Complex neighborhood sampling process. \\
		\hline
		GraphConsis \cite{10.1145/3397271.3401253} & Addresses context, feature, and relation inconsistencies in GNN. & Enhances node representation by resolving inconsistencies. & High complexity in designing and training. \\
		\hline
		Heterogeneous Graph Embeddings \cite{10.14778/3494124.3494128} & Utilizes attention mechanisms and message passing for diverse node types. & Effective in capturing diverse relationships, provides explainability. & Computationally intensive, complex attention mechanisms. \\
		\hline
		Autoencoder \cite{9163376,ebiaredoh2021artificial} & Neural network model for unsupervised learning, reconstructing inputs to detect anomalies. & Effective for anomaly detection, dimensionality reduction. & May suffer from overfitting, requires careful tuning. \\
		\hline
		GEM \cite{10.1145/3269206.3272010} & Heterogeneous graph-based algorithm for malicious account detection. & Learns discriminative embeddings, utilizes attention mechanisms. & Complexity in implementation and training, may require extensive computational resources. \\
		\hline
		\caption{Comparison of various models and techniques for fraud detection in credit card transactions.}\label{table:lit}
	
	\end{longtable}

While the methods discussed above claim to perform well with unbalanced heterogeneous graph datasets, techniques such as autoencoders and decoders, as presented in \cite{9163376,ebiaredoh2021artificial,ebiaredoh2020integrating}, offer alternative solutions. For instance, \cite{ebiaredoh2020integrating} successfully addressed imbalanced medical datasets using a modified Sparse Autoencoder (SAE) and Softmax regression for enhanced diagnosis. However, SAEs are less suitable for data with inherent relationships between elements, which is particularly relevant for fraud detection in transactional networks, where connections between nodes are crucial for identifying suspicious activity. Similarly, \cite{ebiaredoh2021artificial} employed a Stacked SAE (SSAE) for credit card default prediction on imbalanced data. Nevertheless, SSAEs, like SAEs, lack the ability to explicitly prioritize information from relevant neighboring nodes. This limitation necessitates a different approach for this work, which leverages a transactional network to represent data and identify fraudulent activities.

\section{Problem Statement}
\label{ps}
A heterogeneous graph is a specialized graph data structure that comprises multiple types of nodes and edges, wherein each node or edge is uniquely associated with a distinct type. In essence, it represents a graph in which diverse node and edge types are interconnected. To provide a formal definition, the characteristics of a heterogeneous graph are delineated as follows:

\begin{definition}
A heterogeneous graph, also known as a heterogeneous information network or heterogeneous network, is mathematically defined as $G = (V, E, T, R, X)$, where:

\begin{itemize}
    \item $V$ represents the set of nodes in the graph, and each node $v^{t} \in V$ is associated with a specific type $t\in T$, where $T$ represents the set of node types.
    \item $E$ represents the set of edges in the graph, and each edge $e^{r} \in E$ connects two nodes $(v^{t_1}, v^{t_2})$, where $t_1$ and $t_2$ are node types, and $r \in R$, where $R$ represents the set of edge types or relationships.
    \item $X = \{X_v, X_e\}$ represents attributes of nodes and edges, respectively, where $X_v$ represents the set of node attributes, and each node $v^t \in V$ can have a vector of attributes $x_{v^t}$, and $X_e$ represents the set of edge attributes, where each edge $e^r \in E$ can have a vector of attributes $x_{e^r}$.
\end{itemize}

\end{definition}
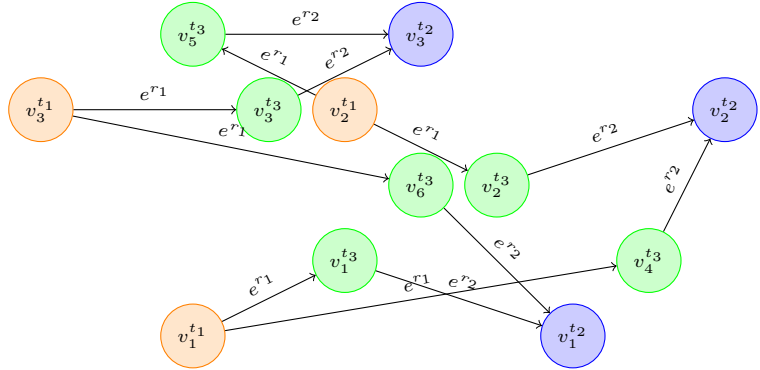
\begin{figure}[ht!]
	\centering
	\begin{tikzpicture}[
		every node/.style={font=\footnotesize, align=center},
		customer/.style={circle, draw=orange, fill=orange!20},
		merchant/.style={circle, draw=blue, fill=blue!20},
		transaction/.style={circle, draw=green, fill=green!20}
		]
		
		\node[customer] (c1) at (0,0) {$v_1^{t_1}$};
		\node[customer] (c2) at (2,3) {$v_2^{t_1}$};
		\node[customer] (c3) at (-2,3) {$v_3^{t_1}$};
		
		\node[merchant] (m1) at (5,0) {$v_1^{t_2}$};
		\node[merchant] (m2) at (7,3) {$v_2^{t_2}$};
		\node[merchant] (m3) at (3,4) {$v_3^{t_2}$};
		
		\node[transaction] (t1) at (2,1) {$v_1^{t_3}$};
		\node[transaction] (t2) at (4,2) {$v_2^{t_3}$};
		\node[transaction] (t3) at (1,3) {$v_3^{t_3}$};
		\node[transaction] (t4) at (6,1) {$v_4^{t_3}$};
		\node[transaction] (t5) at (0,4) {$v_5^{t_3}$};
		\node[transaction] (t6) at (3,2) {$v_6^{t_3}$};
		
		\draw[->] (c1) -- node[above, sloped] {$e^{r_1}$} (t1);
		\draw[->] (t1) -- node[above, sloped] {$e^{r_2}$} (m1);
		
		\draw[->] (c2) -- node[above, sloped] {$e^{r_1}$} (t2);
		\draw[->] (t2) -- node[above, sloped] {$e^{r_2}$} (m2);
		
		\draw[->] (c3) -- node[above, sloped] {$e^{r_1}$} (t3);
		\draw[->] (t3) -- node[above, sloped] {$e^{r_2}$} (m3);
		
		\draw[->] (c1) -- node[above, sloped] {$e^{r_1}$} (t4);
		\draw[->] (t4) -- node[above, sloped] {$e^{r_2}$} (m2);
		
		\draw[->] (c2) -- node[above, sloped] {$e^{r_1}$} (t5);
		\draw[->] (t5) -- node[above, sloped] {$e^{r_2}$} (m3);
		
		\draw[->] (c3) -- node[above, sloped] {$e^{r_1}$} (t6);
		\draw[->] (t6) -- node[above, sloped] {$e^{r_2}$} (m1);
		
	\end{tikzpicture}
	\caption{A heterogeneous graph illustrating different types of nodes and edges.}
	\label{fig:heterogeneous_graph}
\end{figure}

By adhering to its definition, the financial fraud dataset can be depicted as a heterogeneous graph. These datasets encompass various entities, including customer or credit card numbers, merchants' names, and transaction numbers. These entities are represented as nodes in the graph, denoted by $V$. Specifically, the nodes $v^{t_1}$, representing `customer', $v^{t_2}$, representing `merchants', and $v^{t_3}$, representing `transaction', encapsulate the essence of this heterogeneous graph. Consequently, these nodes ($v^{t_1}$, $v^{t_2}$, and $v^{t_3}$) are distinguished by their respective types.

The heterogeneous graph depicted in Figure \ref{fig:heterogeneous_graph} illustrates a network where nodes are categorized into three distinct types: `customers' (in orange), `merchants' (in blue), and `transactions' (in green). Each node type is uniquely identified by an index (\(v_i^{t}\)), where \(i\) indexes different instances of customers, merchants, and transactions within the same type \(t\) (e.g., \(v_1^{t_1}\) for the first customer, \(v_2^{t_1}\) for the second customer). The graph captures complex interactions: customers initiate transactions (\(e^{r_1}\)) that involve merchants (\(e^{r_2}\)). Notably, customers can engage in multiple transactions across different merchants, as represented by multiple transaction nodes (\(v_1^{t_3}, v_2^{t_3}, \ldots, v_6^{t_3}\)). This structured representation facilitates the analysis of interconnected relationships within heterogeneous networks, essential for understanding dynamics financial transactions.\\

\begin{problem}
For the given graph $G = (V, E, T, R, X)$, the task is to determine whether it can be classified as fraudulent, considering that the transaction associated with the graph represents a fraudulent class.
\end{problem}

\section{Methodology} \label{methodology}
\begin{figure}[h]
\centering
\includegraphics[width=\textwidth]{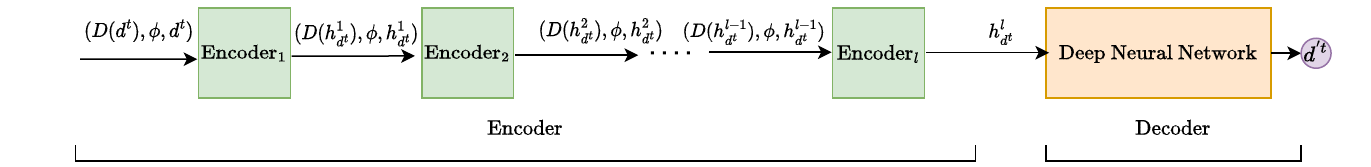}
\caption{Encoder Units and Decoder unit of the model}
\label{fig:model_1}
\end{figure}

The primary objective of this paper is to develop an encoder capable of learning graph embeddings for a given graph \(G = (V, E, T, R, X)\). This encoder will be specifically designed to effectively capture the complex information present in a heterogeneous graph, including both its structure and its attributes. Subsequently, a decoder function \(f_{\text{dec}}\) will be introduced to reconstruct the graph. Figure \ref{fig:model_1} defines the model to be used in this paper.

In this model, there are \(l\) encoder units. The first encoder unit takes (\(D(d^t), \phi, d^t\)) as input, where \(D(d^t)\) represents the source nodes of \(d^t\in V\), and \(\phi\) represents edges \(e^r\) for each source node to \(d^t\). Each encoder unit processes these inputs to produce intermediate representations. The final output of \(Encoder_l\) is fed into a decoder unit, which is implemented as a deep neural network. This decoder unit utilizes the encoded information to generate \(d^{'t}\).  

Finally, the model will calculate the reconstruction error by comparing the reconstructed graph and the original graph. This error serves as a measure of the dissimilarity between the original input and the reconstructed output. Using this error, a threshold for the reconstruction error is established to identify data points that deviate significantly from the  normal patterns. Any data point with a reconstruction error that exceeds the threshold is classified as an anomaly, indicating a deviation from expected normal behavior.

\subsection{Encoder for Heterogeneous Graph}
\begin{figure}[h]
\centering
\includegraphics[width=\textwidth]{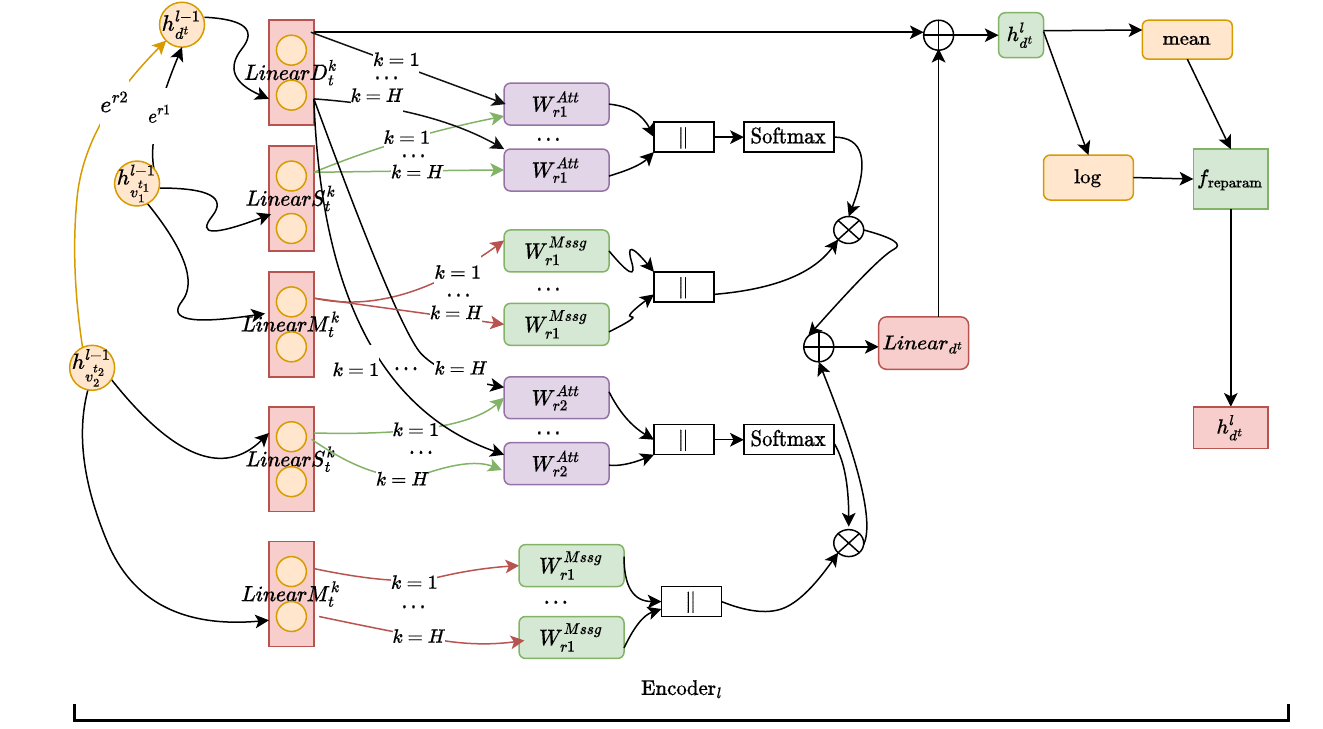}
\caption{Encoder Unit for Heterogeneous Graphs. $e^{r1}$ and $e^{r2}$ denote edges from source nodes $v_1^{t_1}$ and $v_2^{t_2}$ to destination node $d^t$. At $l=0$, it represents the initial encoder layer, producing $h^1_{d^t}$ and so on. $k$ ranges from $1$ to $H$ and $|$ signifies concentration, $\oplus$ denotes addition, and $\otimes$ indicates dot product.}
\label{fig:model_2}
\end{figure}

Based on the study by (\cite{10.1145/3366423.3380027}), a heterogeneous graph encoder for the auto-encoder has been designed (Figure \ref{fig:model_2}). For each destination node $d^t \in V$ and $D(d^t) \in V$, which represents a list of source nodes for $d^t$, the encoding process $f^{\text{enc}}$ is applied as follows:

\begin{equation}
h^{l}_{d^t} = f_{reparam}\Big(\text{Linear}_{d^t} \big( f^{\text{enc}}_{\forall v \in D(d^{t})}(h^{l-1}_{v^t}, e^r, h^{l-1}_{d^t}) \big) \oplus h^{0}_{d^t},mean(h^{l-1}_{d^t}), log(h^{l-1}_{d^t}) \Big)
\label{equa_01}
\end{equation}
\\
Here, $l = 1, 2, \ldots, E_{L}$ represents the encoder layer with a maximum of $E_{L}$ layers, and the initial values are set as $(h^{0}{v^t}, e^r, h^{0}{d^t}) = (v^t, e^r, d^t)$. Additionally, $Linear_{d^t} : \mathbb{R}^{\frac{dim}{k}} \rightarrow \mathbb{R}^{dim}$ denotes the linear projection.

The encoding process $f^{enc}$ can be broken down as:

\begin{equation}
f^{\text{enc}}_{\forall v^t \in D(d^{t})} = \underset{\forall v^t \in D(d^{t})}{\oplus} \Big(f^{Attent}(v^t, e^r, d^t) \cdot f^{Mssg}(v^t, e^r, d^t) \Big)
\label{equ:f_encode}
\end{equation}

\noindent In Equation (\ref{equ:f_encode}), the graph attention mechanism is used and the graph message to embed the node feature of the descriptor node $d_t$ on edges $e^r$. The attention mechanism is defined as follows:

\begin{equation}
f^{Attent}(v^t, e^r, d^t) = \text{Softmax} \Big( \underset{\forall k \in [1,H]}{\|Att^k(v^t, e^r, d^t ) \Big) }
\label{equ:attnd}
\end{equation}

\noindent Inspired by \cite{NIPS2017_3f5ee243}, the attention for each edge $e^r$ is calculated using $k$-heads, based on the dot product of the linear projection of $v \in D(d^t)$ and $d^t$, with a matrix $W^{Att}_{r}$ that depends on the edge, as shown in Equation (\ref{equ:dot_ptod_att}):

\begin{equation}
Att^k(v^t, e^r, d^t) = LinearS_{t}^{k} \Big( h^{l-1}_{v^{t}} \Big) \cdot W^{Att}_{r} \cdot \Big( LinearD_{t}^{k} \Big( h^{l-1}_{d^t} \Big) \Big)^T
\label{equ:dot_ptod_att}
\end{equation}

\noindent For each attention head $k$, $LinearS_{t}^{k} \Big( h^{l-1}_{v^{t}} \Big)$ and $LinearD{t}^{k} \Big( h^{l-1}_{d^t} \Big)$ perform linear projections of the source node $v^t$ and $d^t$, respectively, as well as their subsequent embedded forms. These projections map from $\mathbb{R}^{dim}$ to $\mathbb{R}^{\frac{dim}{k}}$, where $\frac{dim}{k}$ represents the vector dimension for each head. Finally, $W^{Att}_{r} \in \mathbb{R}^{\frac{dim}{k} \times \frac{dim}{k}}$ is a learnable edge matrix for the edge type $r$ connecting between $v^t$ and $d^t$.

\noindent The message passing function for each attention head $k$ is obtained by performing the dot product between the linear projection of the source node $v^t$ and a matrix $W^{Mssg}_{r} \in \mathbb{R}^{\frac{dim}{k}}$ based on the edge type joining $v^t$ and $d^t$. The linear projection is carried out by $LinearM^{k}_{t}$ for each type of node, and thus the final projection is mapped from $\mathbb{R}^{dim}$ to $\mathbb{R}^{\frac{dim}{k}}$, as shown in Equation (\ref{equ:mssg}):

\begin{equation}
f^{Mssg}(v^t,e^r,d^t) = \underset{ \forall k \in [1,H]} \| LinearM^{k}{t} \Big( h^{l-1}_{v^t} \Big) W^{Mssg}_{r}
\label{equ:mssg}
\end{equation}

Finally, a function called \( f_{\text{reparam}} \) creates a probabilistic model by remodeling the latent variable \( h^{l}_{d^t} \) using probabilistic distributions, facilitating gradient-based optimization. This approach captures uncertainty and generates diverse samples. Following \cite{Kingma_2019}, the function is written as follows:

	\begin{align*}
		h^{l}_{d^t} &= f_{\text{reparam}}\Big(\text{mean}(h^{l}_{d^t}), \log(h^{l}_{d^t}) \Big)\\
		&= \text{mean}(h^{l}_{d^t}) + \epsilon \cdot \exp\left(\frac{1}{2} \cdot \log(h^{l}_{d^t}) \right)
	\end{align*}

where $\epsilon = \mathcal{N}(0, 1)$ (sampled random noise).

\subsection{Decoder For Heterogeneous Graph}

The graph decoder should account for the heterogeneous nature of the graph $G$ when reconstructing the original structure. It needs to reconstruct the specific types of nodes and edges, ensuring that the reconstructed graph maintains the semantic relationships and attributes associated with each type. This requires incorporating type-specific reconstruction mechanisms into the decoder.

For each node $d^t$, a node decoder $f_{\text{dec}}$ is applied  to reconstruct the attributes $d^{'t}$ based on the corresponding node embedding $h^l_{d^t}$:

\begin{equation}
d^{'t} = f_{\text{dec}}(h^l_{d^t})
\label{equ:dec}
\end{equation}

\noindent Here, $d^{'t}$ represents the reconstructed attribute of node $d^t$ with the original attribute $h^0_{d^t}$.

The primary objective of the decoder is to reconstruct the original graph data, which is based on the graph embeddings generated by the encoder. The overall loss function for the autoencoder will be expressed as follows:

\begin{equation}
L = \sum_{\forall N} \sum_{\forall t} \text{LOSS}(d^t, d^{'t})
\label{equ:loss}
\end{equation}

\noindent The loss function $L$ represents the sum of losses calculated by the $\text{LOSS}$ function between the original graph data and the reconstructed data.

\subsection{Algorithm}

\begin{algorithm}
	\caption{Fraud Detection on a Heterogeneous Graph}
	\label{alg:fraud_detection}
	\begin{algorithmic}[1]
		\Require Heterogeneous Graph $G$
		\Ensure `Fraud' or `Not Fraud'
		\For{$d^t \in G$}
		\State $(h^{0}_{v^t}, e^r, h^{0}_{d^t}) \leftarrow (v^t, e^r, d^t)$ \Comment{Initialization}
		\For{$l \leftarrow 1$ \textbf{to} $E_L$} \Comment{Message Passing Layers}
		\For{$v^t \in D(d^t)$} \Comment{Neighborhood of $d^t$}
		\State $h^{l}_{d^t} = \text{Linear}_{d^t}\big( f^{\text{enc}}(h^{l-1}_{v^t}, e^r, h^{l-1}_{d^t}) \big) \oplus h^{0}_{d^t}$
		\EndFor
		\State $h^{l}_{d^t} = f_{\text{reparam}}\Big(\text{mean}(h^{l}_{d^t}), \log(h^{l}_{d^t})\Big)$ \Comment{Reparameterization}
		\EndFor
		\State $d^{'t} = f_{\text{dec}}(h^{l}_{d^t})$ \Comment{Output Layer}
		\EndFor
		\State $L = \text{LOSS}(d^t, d^{'t})$ \Comment{Loss Calculation}
		\If{$L < \text{Threshold}()$}
		\State \Return `Non-Fraud'
		\Else
		\State \Return `Fraud'
		\EndIf
	\end{algorithmic}
\end{algorithm}

The algorithm depicted in Algorithm~\ref{alg:fraud_detection}, outlines the method for detecting fraud in a heterogeneous graph structure. Here's a detailed breakdown of each step:
\begin{enumerate}
	\item {Input (Heterogeneous Graph G): } This represents the financial transaction network, containing nodes (customers, merchants, transactions) and edges (interactions) with their respective types.
	\item {Output:} \textbf{``Fraud'' or ``Not Fraud''}: The algorithm classifies the transaction associated with the input graph as either fraudulent or legitimate.
	\item {Algorithm Steps: }
	\begin{itemize}
		\item For each node \( d^t \) in the graph \( G \), node \( d^t \) is initialized with \( (h^{0}_{v^t}, e^r, h^{0}_{d^t}) \). It includes the features of the node itself \( h^{0}_{d^t} \), the connecting edge type \( e^r \), and the initial representation of the source node \( h^{0}_{v^t} \).
		\item {Message Passing Layers (L Layers):}
		\begin{itemize}
				\item This loop iterates through a predefined number of layers (\( E_L \)) in the GNN architecture.
			\item Within each layer \( l \):
			\begin{itemize}
				\item For each node \( v^t \) in the neighborhood of the current node \( d^t \):
				\begin{itemize}
					\item 	A message function \( f^{\text{enc}} \) (Equation \ref{equ:f_encode}) aggregates information from the source node’s hidden representation \( h^{l-1}_{v^t} \), the edge type \( e^r \), and the previous hidden representation of the destination node \( h^{l-1}_{d^t} \). The message undergoes a linear transformation with \( \text{Linear}_{d^t} \) as per equations (\ref{equ:attnd}-\ref{equ:mssg}).
					
					\item By utilizing the attention mechanism, the messages undergo transformation and are subsequently combined with the initial hidden representation of the destination node \( h^{0}_{d^t} \) through element-wise addition (\( \oplus \)).
				\end{itemize}
				\item The message passing happens iteratively for all neighbors of \( d^t \).
				\item The updated hidden representation \( h^{l}_{d^t} \) is subjected to \( f_{\text{reparam}} \) (Equation \ref{equa_01}) after message aggregation. Mean and logarithm are utilized in hidden representation to ensure greater stability during training.
		\end{itemize}
		
	\end{itemize}
\end{itemize}
		\item {Output Layer:} The final hidden representation \( h^{l}_{d^t} \) is passed through the decoder function \( f_{\text{dec}} \) (Equation \ref{equ:dec}) to produce the prediction vector \( d^{’t} \).
		\item {Loss Calculation:}  The difference between the predicted output \( d^{’t} \) and the original node feature \( d^{t} \) is evaluated using a loss function \( \text{LOSS} \). The LOSS function can use a metric such as mean squared error or any other appropriate loss function.
		
		\item {Fraud Classification:}  A threshold function \( \text{Threshold}() \) is used to determine the classification based on the calculated loss. If the loss is lower than the threshold (indicating a good fit), the algorithm outputs ``Non-Fraud''. Conversely, if the loss is higher than the threshold (indicating a poor fit), it outputs ``Fraud''.
		
\end{enumerate}

Algorithm \ref{alg:fraud_detection} explains the entire framework of the model, which is designed to identify if a specific data point is linked to fraudulent behavior, resulting in one of two possible outcomes: `Fraud' or `Not Fraud.' The algorithm calculates a loss value to measure the difference between the original transaction node and its decoded version. The computation of this loss relies on a loss function that has been predetermined. The next step in the process is for the algorithm to compare the resulting loss with a predetermined threshold, once all the calculations have been completed. In the case where the loss falls below the designated threshold, the data point is classified as ‘Not Fraud’. The overall time complexity of the algorithm can be approximated as $\mathcal{O}(nE)$ by summing up these components, with $n$ representing the number of nodes in the graph.

\section{Experiment} \label{imp-res}

This paper assesses the effectiveness of the proposed model through a series of experiments on credit card fraud datasets and a comparison with other existing machine learning and deep learning models.

\subsection{Performance Metrics}
In order to evaluate the performance of various models, this article employed evaluation metrics that include the precision rate (PR), the recall rate (RR), the ROC curve, and the F1 score. These metrics are defined as follows: 

$$ \text{PR} = \frac{\text{TP}}{\text{TP} + \text{FP}}$$
$$\text{RR} = \frac{\text{TP}}{\text{TP} + \text{FN}}$$

In this context, true positive (TP) and false positive (FP) indicate the number of correctly and incorrectly predicted instances of fraud, respectively. Conversely, true negative (TN) and false negative (FN) correspond to the count of transactions accurately and inaccurately predicted as non-fraudulent.

Meanwhile, the ROC curve illustrates the classifier's ability to differentiate between fraud and non-fraud categories. This curve is created by plotting the true positive rate against the false positive rate at different threshold levels. The AUC, which ranges from 0 to 1, encapsulates the information from the ROC curve. A value of 0 signifies that all classifier predictions are erroneous, while a value of 1 indicates a perfect classifier.

The F1 score represents the harmonic mean of precision and recall. Precision is the ratio of true positive predictions to the total predicted positives and recall is the ratio of true positive predictions to the total actual positives. It provides a single value that harmonizes precision and recall, facilitating a balanced evaluation of classifier performance.

$$\text{F1} = 2 * \frac{(\text{PR} * \text{RR})}{(\text{PR} + \text{RR})}$$

Given that the dataset is imbalanced, the F1 score is particularly valuable because it considers both precision and recall. This score provides a straightforward way to assess a classifier's overall effectiveness in accurately identifying positive instances while minimizing false positives and false negatives.

Another parameter used to gain insight into the model's performance is the Precision-Recall curve (AUC-PR) \cite{Powers2011EvaluationFP}. This metric offers valuable insights, particularly in situations where class distribution is imbalanced \cite{neptune2023}.

\subsection{Datasets}
The dataset  (\cite{Kartik2021}) used in this article simulates credit card transactions and includes genuine and fraudulent activities that occurred between January 1, 2019, and December 31, 2020. The data encompass transactions carried out by 1000 customers using credit cards issued by a variety of banks, engaging in transactions with a pool of 800 different merchants.

\begin{table}[ht]
  \centering
  
  \begin{tabular}{|l|c|c|}
    \toprule
    \textbf{Types of Dataset} & \textbf{Normal Data} & \textbf{Abnormal Data} \\
    \midrule
    Training Dataset & 1842743 & 9651 \\
    \hline
    Testing Dataset & 553574 & 2145 \\
    \bottomrule
  \end{tabular}
 
  \caption{Distribution of Fraudulent Transactions on Training and Testing Dataset}
   \label{tab:dataset_types}
  \end{table}
Table \ref{tab:dataset_types} illustrates the distribution of fraudulent and non-fraudulent transactions in a dataset. It shows the number of occurrences of each type of transaction, with ``1" representing fraudulent (Abnormal) transactions and ``0" representing non-fraudulent (Normal) ones. This analysis gives an indication of the skewed and unbalanced ratio of fraudulent to non-fraudulent transactions.

\subsection{Analysis of Algorithms}

In the article (\cite{Arora2020}), some of the best machine learning algorithms that handle fraud datasets are listed. Here is the list used in the article:

\begin{itemize}
\item Linear Regression
\item Logistic Regression
\item Decision Tree
\item SVM (Support Vector Machine)
\item ANN (Artificial Neural Network)
\item Naïve Bayes
\item DNN (Deep Neural Network)
\item K-Means
\item Random Forest
\item Dimensionality Reduction Algorithms
\item Gradient Boosting (XGB) Algorithms
\end{itemize}

These algorithms cover a wide range of machine learning (ML) aspects, including association analysis, clustering, classification, statistical learning, and link mining. They hold a crucial place among the essential topics explored in research and development within the field of machine learning. However, when evaluating these algorithms with datasets, their performance often falls short of expectations due to the inherent imbalance present in the data.

Table \ref{tab:result_01} provides the performance metrics for a few machine learning algorithms. The table shows that the F1 score of all machine learning algorithms is too low, suggesting that these algorithms could not handle unbalanced datasets properly. Since the F1 score is low and the AUC curve is high for all ML algorithms, it indicates that these algorithms are adept at distinguishing between abnormal and normal data, as evidenced by the high AUC value. However, the F1 score is low due to the models facing challenges in achieving both high precision and high recall, attributed to the imbalanced nature of the data.
\begin{table}[h!]
    \centering
\begin{tabular}{ |p{3 cm}||p{1.5cm}|p{1.5cm}|p{1.5cm}|p{1cm}|p{1 cm}|  }
 \hline
 \multicolumn{6}{|c|}{Performance of Machine learning (ML) Algorithms} \\
 \hline
 ML Algorithm &Testing Accuracy&Testing F1 Score&Test Precision &Test Recall &AUC\\
 \hline
 Decision Tree   & $0.99$ &$0.29$&$0.22$&$0.43$& $0.82$\\
 \hline
 XGB Classifier  & $0.99$ &$0.33$&$0.27$&$0.43$& $\textbf{0.96}$\\
 \hline
 ANN  & $0.99$ &$0.33$&$0.23$&$0.32$& $0.92$\\
 \hline
 Deep NN  &$0.99$&$0.33$&$\textbf{0.40}$&$0.26$& $0.81$\\
 \hline
 AE & $-$ &$\textbf{0.67}$&$0.50$&$0.99$& $0.52$\\
 \hline
  VAE & $-$ &$\textbf{0.67}$&$0.50$&$0.99$& $0.54$\\
  \hline
  Sparse AE & $-$ &$\textbf{0.67}$&$0.50$&$0.99$& $0.54$\\
 
 \hline
\end{tabular}
    \caption{Performance Measurement of Few Selected Machine Learning Algorithms}
    \label{tab:result_01}
\end{table}

These scenarios arise when the negative class dominates the dataset, creating a highly imbalanced situation. In such cases, models tend to classify instances as the majority class, resulting in high true negatives and low false positives but at the cost of missing true positives and having low recall. To address the challenges posed by unbalanced data, various algorithms are explored. One of the algorithms under consideration is the autoencoder algorithm. 

The exploration involves simple autoencoders (AE) using deep neural networks and their variations, such as variational autoencoders (VAE) and sparse autoencoders (Sparse AE). Table \ref{tab:result_01} also shows the performance of the autoencoders. Regardless of the specific type, the model's performance is evaluated using key metrics. The F1 score, which harmonizes precision and recall, yielded a value of 0.67. This suggests that the models have achieved a reasonable balance between making accurate positive predictions and effectively capturing actual positive instances. Overall, the performance is decent, showing a well-rounded approach.

However, the narrative changes when examining the Receiver Operating Characteristics (ROC) curve and its corresponding Area Under the Curve (AUC). With an AUC of 0.57, it implies that the models struggle to distinguish between fraud and normal classes. Their ability to classify effectively in this context appears limited and performs only slightly better than random guessing.

In a deeper dive, the precision achieved by the autoencoder models in the test set is 0.50. This means that roughly half of the abnormal predictions it generates are accurate, while the other half are incorrect. On the other hand, the recall rate is impressive at 0.99. This means that the models excel at identifying almost all the actual abnormal instances present in the dataset.

In summary, while autoencoder models demonstrate balanced performance in terms of the F1 score, with commendable recall and reasonable precision, the AUC score and precision rates indicate room for improvement. Enhancing the discriminatory capacity of models and refining their positive prediction accuracy could be areas of focus to further elevate their performance in classification tasks.

\subsection{Analysis of the Proposed Model}

\begin{table}[h!]
    \centering
    \begin{tabular}{ |p{6cm}|p{1.5cm}|  }
        \hline
        \textbf{Parameter Name} & \textbf{Value}\\
        \hline
        Size of Hidden Layers & $64$ \\
        \hline
        Number of heads ($H$) & $16$ \\
        \hline
        Number of Layers for the Encoder ($l$) & $124$ \\
        \hline
        Number of Layers for the Decoder & $64$\\
        \hline
        Dropout Rate & $0.4$ \\
        \hline
        Regularization Rate & $0.01$\\
        \hline
    \end{tabular}
    \caption{Values for different parameters used in the model.}
    \label{tab:result_02}
\end{table}

\begin{figure}[ht!]
	\centering
	\begin{subfigure}[b]{0.45\textwidth}
		\includegraphics[width=\textwidth]{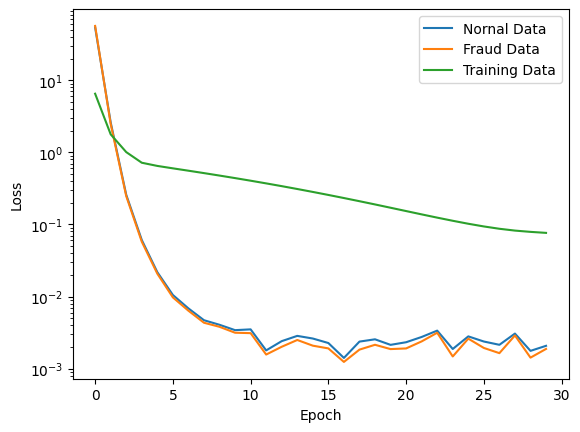}
		\caption{Training loss and evaluation loss.}
		\label{fig:modelLoss}
	\end{subfigure}
	\hfill
	\begin{subfigure}[b]{0.45\textwidth}
		\includegraphics[width=\textwidth]{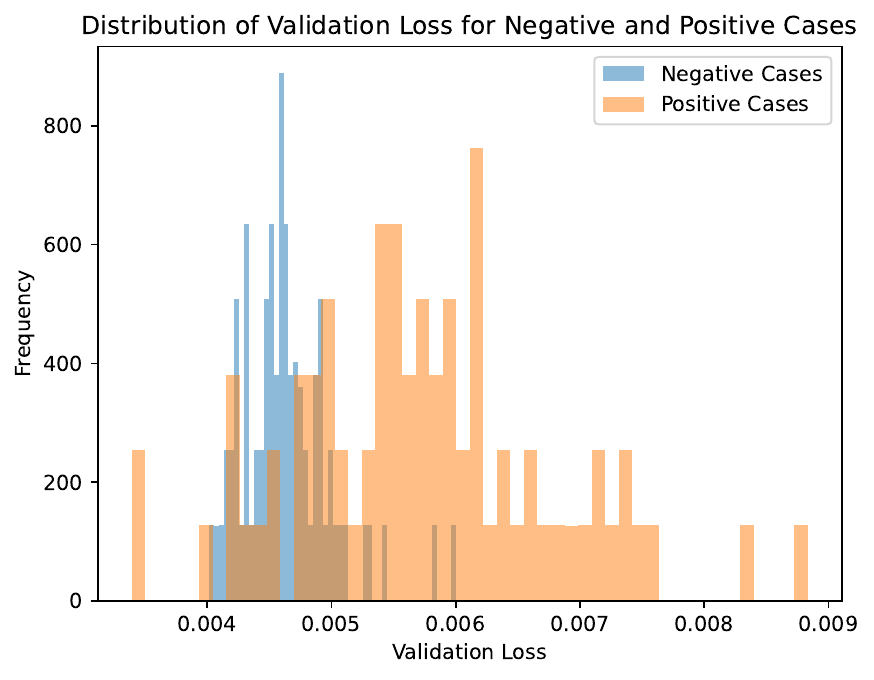}
		\caption{Distribution of validation loss.}
		\label{fig:DistValidLoss}
	\end{subfigure}
	
	\vspace{1em} 
	
	\begin{subfigure}[b]{0.45\textwidth}
		\includegraphics[width=\textwidth]{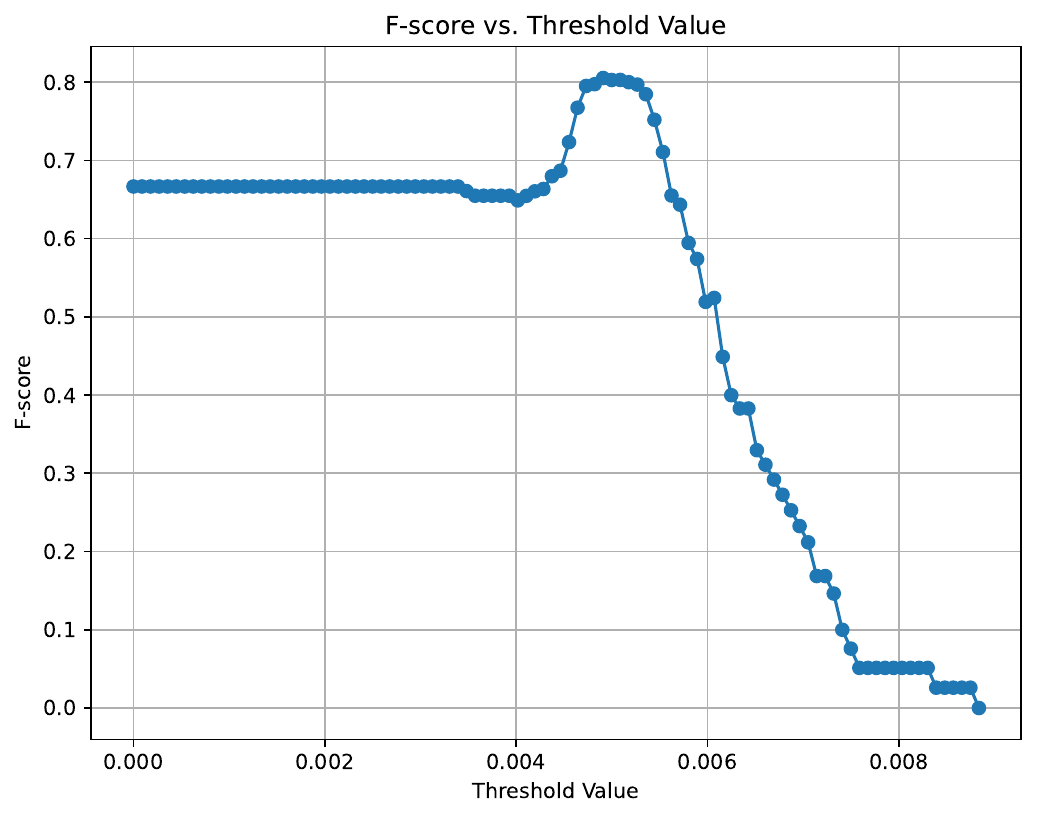}
		\caption{F Score vs Threshold graph.}
		\label{fig:fscore}
	\end{subfigure}
	\hfill
	\begin{subfigure}[b]{0.45\textwidth}
		\includegraphics[width=\textwidth]{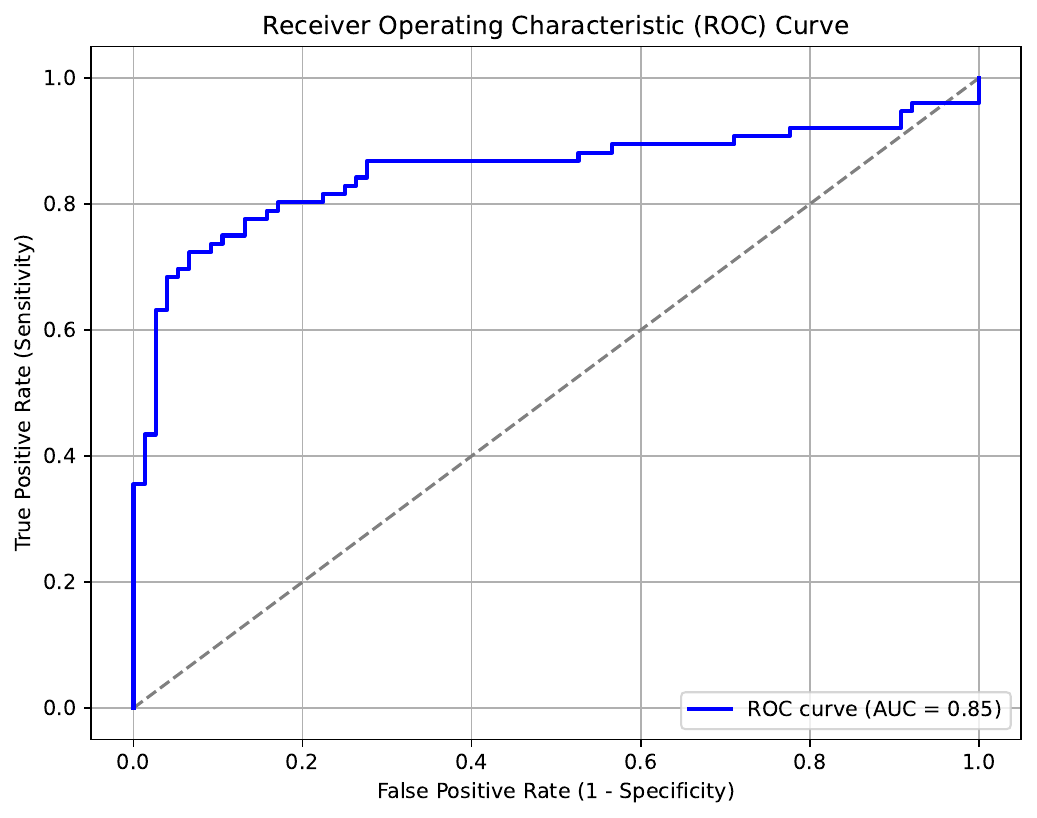}
		\caption{ROC curve}
		\label{fig:roccurve}
	\end{subfigure}
	\hfill
	\begin{subfigure}[b]{0.45\textwidth}
		\centering
		\includegraphics[width=\textwidth]{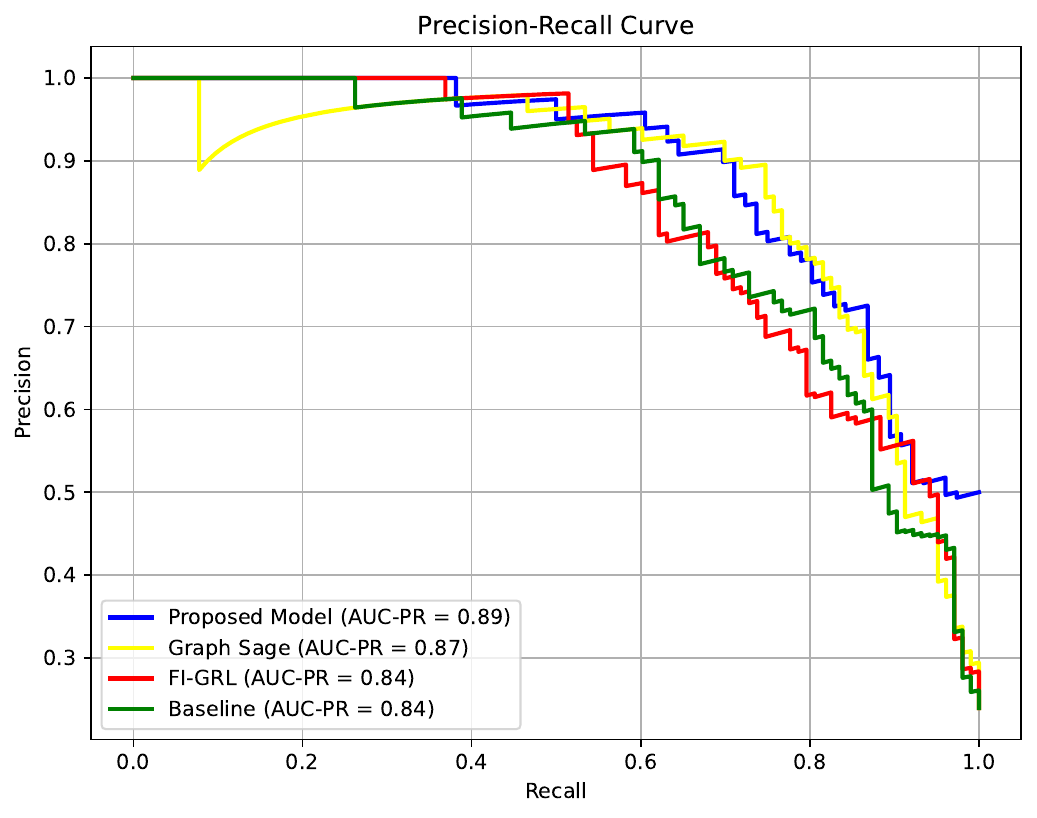}
		\caption{Precision-Recall curve of the model with an AUC-PR of 0.89.}
		\label{fig:AUC_PR}
	\end{subfigure}
	
	\caption{Performance evaluations of the proposed model.}
	\label{fig:perfom}
\end{figure}
After tuning the parameters for different hyperparameters, the performance of the model is represented as shown in Figure~\ref{fig:perfom}. Finally, the proposed model uses the parameters defined in Table~\ref{tab:result_02} to evaluate the model's performance.

In Figure~\ref{fig:modelLoss}, the training loss is compared with the validation loss for positive (fraud) and negative datasets. This plot provides insight into how effectively the model handles overfitting and underfitting of the data. The model, using the parameters from Table~\ref{tab:result_02}, demonstrates immunity to both overfitting and underfitting, effectively managing these issues. Figure~\ref{fig:DistValidLoss} illustrates the loss distribution (histogram) generated by the model from the dataset. This distribution shows the loss values for both positive and negative data in the dataset. The figure reveals that the loss for negative instances is concentrated between $0.004$ and $0.005$, while the loss for positive instances is distributed beyond $0.006$. 

Figure~\ref{fig:fscore} defines the model's F1 score versus the classification threshold value. From the figure, it can be seen that the F1 score reaches its highest value of 0.81 at a loss value of 0.005. Additionally, the ROC curve was plotted based on the threshold, resulting in the ROC curve shown in Figure~\ref{fig:roccurve}, and an AUC of 0.85 was obtained for the model.

The Precision-Recall (PR) curve (Figure \ref{fig:AUC_PR}) compares the performance of four algorithms: the Proposed Model, Graph Sage \cite{VANBELLE2022116463}, FI-GRL \cite{VANBELLE2022116463}, and Baseline \cite{VANBELLE2022116463}. The Proposed Model exhibits the highest performance with an AUC-PR of 0.89, indicating the best balance between precision and recall. Graph Sage follows closely with an AUC-PR of 0.87, showing strong but slightly inferior performance compared to the Proposed Model. Both FI-GRL and the Baseline models have an AUC-PR of 0.84, indicating moderate performance and similar effectiveness in maintaining precision and recall. Overall, the Proposed Model stands out as the most effective, followed by Graph Sage, with FI-GRL and Baseline performing similarly but less effectively.

Again, Table \ref{tab:result_02} summarizes the performance of various graph learning algorithms on metrics including AUC-PR, F1-Score, and ROC-AUC. The proposed model achieves the highest AUC-PR (0.89) and F1-Score (0.81) but has a lower ROC-AUC (0.85) compared to Graph Sage and XBoost, which achieve a ROC-AUC of 0.93.
\begin{table}[h!]
    \centering
\begin{tabular}{ |p{3 cm}|p{2 cm}|p{2 cm}|p {2 cm}|  }
 \hline
 \multicolumn{4}{|c|}{Performance of Graph Learning Algotihms} \\
 \hline
 \textbf{Graph Algorithm} & \textbf{AUC-PR} & \textbf{F1 Score} & \textbf{ROC-AUC}\\
 \hline
 Proposed Model   & \textbf{0.89}&\textbf{0.81}& 0.85\\
 \hline
 Graph Sage and XBoost (\cite{VANBELLE2022116463}) &0.86&0.80&\textbf{0.93}\\
 \hline
 FI-GRL(\cite{VANBELLE2022116463})&0.84&$0.70$&$0.92$\\
  \hline
 Baseline(\cite{VANBELLE2022116463})&0.84&$0.74$&$0.91$\\
 
 \hline
\end{tabular}
    \caption{Performance Measurement of Graph Learning Algorithms. AUC-PR provides sufficient information to assess performance due to the imbalanced nature of the dataset used.}
    \label{tab:result_02}
\end{table}

\begin{figure}[h!]
\centering
\includegraphics[width=0.7\textwidth]{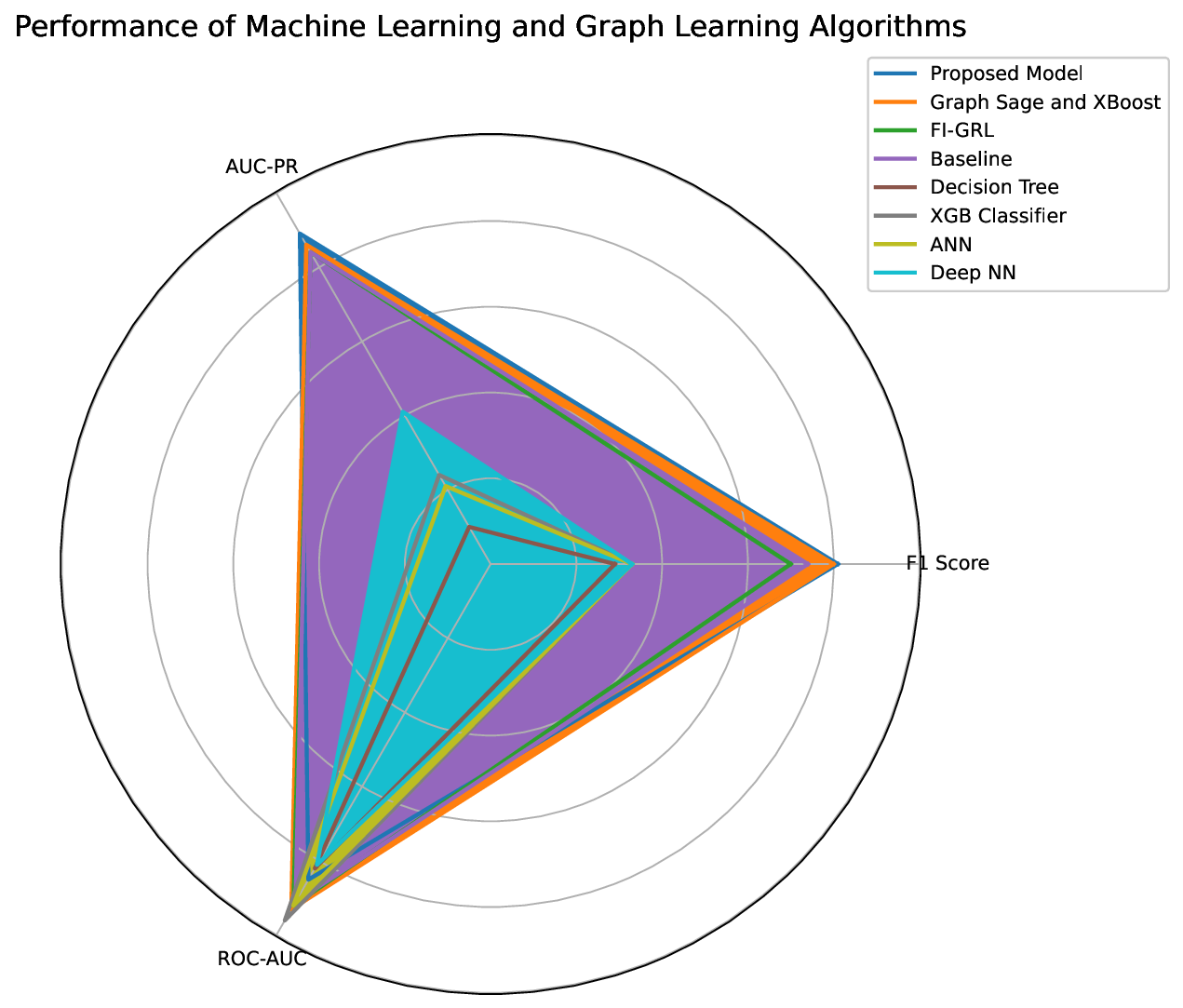}
\caption{Performance Radar Chart. It compares several machine learning (ML) algorithms, including the Proposed Algorithm, using three key metrics: F1 Score, AUC-PR (Area Under the Precision-Recall Curve), and ROC-AUC (Area Under the Receiver Operating Characteristic Curve). This visualization highlights the strengths and weaknesses of each algorithm across these important performance metrics, providing a comprehensive view of their comparative effectiveness.}
\label{fig:radar_chart}
\end{figure}

Finally, Figure \ref{fig:radar_chart} showcases the following algorithms: Proposed Model, Graph Sage and XBoost, FI-GRL, Baseline, Decision Tree, XGB Classifier, ANN (Artificial Neural Network), and Deep NN (Deep Neural Network). This radar chart highlights the exceptional performance of the Proposed Model, with high scores in F1 score, AUC-PR, and ROC-AUC, demonstrating a strong and balanced performance.

While Graph Sage and XBoost show great performance in class discrimination with high ROC-AUC, their AUC-PR is slightly lower, suggesting a trade-off when dealing with imbalanced datasets. Both FI-GRL and Baseline demonstrate strong classification performance with high ROC-AUC, but they may prioritize precision or recall at the expense of balance, resulting in a lower F1 score.

The Decision Tree and XGB Classifier face challenges in their competition, as the Decision Tree exhibits overall weakness, and the XGB Classifier lacks balance despite its strong classification ability. Finally, ANN and Deep NN exhibit moderate performance across all metrics, lacking a clear specialization. With its balanced performance, the Proposed Model stands out as a strong candidate for general use, unlike other algorithms that focus on specific needs.

\section{Conclusion} \label{conclu}

In this paper, a novel approach is introduced that incorporates a heterogeneous graph autoencoder with an attention mechanism, designed to extract valuable information from the intricate graph structure. The encoded node information, produced by the encoder, is harnessed to create a probabilistic distribution using a variational autoencoder model, allowing for the capture of uncertainty and the generation of diverse samples of the embedded nodes. This model effectively addresses the first research question. Subsequently, the output of the encoder undergoes further processing by a deep learning neural network, leading to the regeneration of the original node embeddings. This process significantly enhances the embedded representations of the nodes within the heterogeneous graph. The errors generated by the decoder are carefully observed and recorded, playing a crucial role in classifying fraudulent from non-fraudulent transactions. To facilitate this, a straightforward search algorithm is employed to determine an efficient threshold, effectively addressing the second research question. The work is rigorously benchmarked against a selection of state-of-the-art machine learning algorithms and compared with established methods, such as Graph-Sage and FI-GRL, serving as baselines. Remarkably, the approach consistently demonstrates superior performance, outperforming these baseline methods and thus effectively addressing the third research question. However, the model currently lacks the capability to effectively handle temporal data relationships, which is essential for addressing the dynamic nature of datasets, particularly in the context of fraudulent transactions. This issue will be a focal point for future research and development.
\section*{Statements and Declarations}

\subsection*{Competing Interests}
The authors declare that there are no competing interests associated with this research work.

\subsection*{Funding}
This research did not receive any specific grant from funding agencies in the public, commercial, or not-for-profit sectors.

\subsection*{Informed Consent}
Informed consent was obtained from all individual participants included in the study.

\subsection*{Data Availability}
The datasets generated and/or analyzed during the current study are available in upon reasonable request from the corresponding author.

\bibliography{Final}

\begin{thebibliography}{10}
\expandafter\ifx\csname url\endcsname\relax
  \def\url#1{\burl{#1}}\fi
\expandafter\ifx\csname urlprefix\endcsname\relax\def\urlprefix{URL }\fi
\providecommand{\bibinfo}[2]{#2}
\providecommand{\eprint}[2][]{\url{#2}}
\providecommand{\doi}[1]{\url{https://doi.org/#1}}
\bibcommenthead

\bibitem{Financial2022}
\bibinfo{author}{Ali, A.} \emph{et~al.}
\newblock \bibinfo{title}{Financial fraud detection based on machine learning:
  A systematic literature review}.
\newblock \emph{\bibinfo{journal}{Applied Sciences}}
  \textbf{\bibinfo{volume}{12}}, \bibinfo{pages}{9637} (\bibinfo{year}{2022}).
\newblock \urlprefix\url{http://dx.doi.org/10.3390/app12199637}.

\bibitem{hussain2021fraud}
\bibinfo{author}{Hussain, S.~S.}, \bibinfo{author}{Reddy, E. S.~C.},
  \bibinfo{author}{Akshay, K.~G.} \& \bibinfo{author}{Akanksha, T.}
\newblock \emph{\bibinfo{title}{Fraud detection in credit card transactions
  using svm and random forest algorithms}}, \bibinfo{pages}{1013--1017}
  (\bibinfo{organization}{IEEE}, \bibinfo{year}{2021}).

\bibitem{9540093}
\bibinfo{author}{Jing, R.} \emph{et~al.}
\newblock \emph{\bibinfo{title}{A gnn-based few-shot learning model on the
  credit card fraud detection}}, \bibinfo{pages}{320--323}
  (\bibinfo{year}{2021}).

\bibitem{zhou2020graph}
\bibinfo{author}{Zhou, J.} \emph{et~al.}
\newblock \bibinfo{title}{Graph neural networks: A review of methods and
  applications}.
\newblock \emph{\bibinfo{journal}{AI open}} \textbf{\bibinfo{volume}{1}},
  \bibinfo{pages}{57--81} (\bibinfo{year}{2020}).

\bibitem{Bo2023}
\bibinfo{author}{Bo, D.}
\newblock \emph{\bibinfo{title}{Homogeneous Graph Neural Networks}},
  \bibinfo{pages}{27--59} (\bibinfo{publisher}{Springer International
  Publishing}, \bibinfo{address}{Cham}, \bibinfo{year}{2023}).
\newblock \urlprefix\url{https://doi.org/10.1007/978-3-031-16174-2_3}.

\bibitem{Shi2022}
\bibinfo{author}{Shi, C.}, \bibinfo{author}{Wang, X.} \& \bibinfo{author}{Yu,
  P.~S.}
\newblock \emph{\bibinfo{title}{Structure-Preserved Heterogeneous Graph
  Representation}}, \bibinfo{pages}{29--69} (\bibinfo{publisher}{Springer
  Singapore}, \bibinfo{address}{Singapore}, \bibinfo{year}{2022}).
\newblock \urlprefix\url{https://doi.org/10.1007/978-981-16-6166-2_3}.

\bibitem{ijfs11030110}
\bibinfo{author}{Cheah, P. C.~Y.}, \bibinfo{author}{Yang, Y.} \&
  \bibinfo{author}{Lee, B.~G.}
\newblock \bibinfo{title}{Enhancing financial fraud detection through
  addressing class imbalance using hybrid smote-gan techniques}.
\newblock \emph{\bibinfo{journal}{International Journal of Financial Studies}}
  \textbf{\bibinfo{volume}{11}} (\bibinfo{year}{2023}).
\newblock \urlprefix\url{https://www.mdpi.com/2227-7072/11/3/110}.

\bibitem{brownlee2021combine}
\bibinfo{author}{Brownlee, J.}
\newblock \bibinfo{title}{How to combine oversampling and undersampling for
  imbalanced classification}.
\newblock
  \bibinfo{howpublished}{\url{https://machinelearningmastery.com/combine-oversampling-and-undersampling-for-imbalanced-classification/}}
  (\bibinfo{year}{2021}).
\newblock \bibinfo{note}{Accessed: October, 2023}.

\bibitem{4358713}
\bibinfo{author}{Srivastava, A.}, \bibinfo{author}{Kundu, A.},
  \bibinfo{author}{Sural, S.} \& \bibinfo{author}{Majumdar, A.}
\newblock \bibinfo{title}{Credit card fraud detection using hidden markov
  model}.
\newblock \emph{\bibinfo{journal}{IEEE Transactions on Dependable and Secure
  Computing}} \textbf{\bibinfo{volume}{5}}, \bibinfo{pages}{37--48}
  (\bibinfo{year}{2008}).

\bibitem{ROBINSON2018235}
\bibinfo{author}{Robinson, W.~N.} \& \bibinfo{author}{Aria, A.}
\newblock \bibinfo{title}{Sequential fraud detection for prepaid cards using
  hidden markov model divergence}.
\newblock \emph{\bibinfo{journal}{Expert Systems with Applications}}
  \textbf{\bibinfo{volume}{91}}, \bibinfo{pages}{235--251}
  (\bibinfo{year}{2018}).
\newblock
  \urlprefix\url{https://www.sciencedirect.com/science/article/pii/S0957417417305894}.

\bibitem{LUCAS2020393}
\bibinfo{author}{Lucas, Y.} \emph{et~al.}
\newblock \bibinfo{title}{Towards automated feature engineering for credit card
  fraud detection using multi-perspective hmms}.
\newblock \emph{\bibinfo{journal}{Future Generation Computer Systems}}
  \textbf{\bibinfo{volume}{102}}, \bibinfo{pages}{393--402}
  (\bibinfo{year}{2020}).
\newblock
  \urlprefix\url{https://www.sciencedirect.com/science/article/pii/S0167739X19300664}.

\bibitem{Itoo2021}
\bibinfo{author}{Itoo, F.}, \bibinfo{author}{{Meenakshi}} \&
  \bibinfo{author}{Singh, S.}
\newblock \bibinfo{title}{Comparison and analysis of logistic regression,
  na{\"i}ve bayes and knn machine learning algorithms for credit card fraud
  detection}.
\newblock \emph{\bibinfo{journal}{International Journal of Information
  Technology}} \textbf{\bibinfo{volume}{13}}, \bibinfo{pages}{1503--1511}
  (\bibinfo{year}{2021}).
\newblock \urlprefix\url{https://doi.org/10.1007/s41870-020-00430-y}.

\bibitem{9640631}
\bibinfo{author}{Saddam~Hussain, S.~K.}, \bibinfo{author}{Sai Charan~Reddy,
  E.}, \bibinfo{author}{Akshay, K.~G.} \& \bibinfo{author}{Akanksha, T.}
\newblock \emph{\bibinfo{title}{Fraud detection in credit card transactions
  using svm and random forest algorithms}}, \bibinfo{pages}{1013--1017}
  (\bibinfo{year}{2021}).

\bibitem{math9212683}
\bibinfo{author}{Lin, T.-H.} \& \bibinfo{author}{Jiang, J.-R.}
\newblock \bibinfo{title}{Credit card fraud detection with autoencoder and
  probabilistic random forest}.
\newblock \emph{\bibinfo{journal}{Mathematics}} \textbf{\bibinfo{volume}{9}}
  (\bibinfo{year}{2021}).
\newblock \urlprefix\url{https://www.mdpi.com/2227-7390/9/21/2683}.

\bibitem{8292883}
\bibinfo{author}{Randhawa, K.}, \bibinfo{author}{Loo, C.~K.},
  \bibinfo{author}{Seera, M.}, \bibinfo{author}{Lim, C.~P.} \&
  \bibinfo{author}{Nandi, A.~K.}
\newblock \bibinfo{title}{Credit card fraud detection using adaboost and
  majority voting}.
\newblock \emph{\bibinfo{journal}{IEEE Access}} \textbf{\bibinfo{volume}{6}},
  \bibinfo{pages}{14277--14284} (\bibinfo{year}{2018}).

\bibitem{RB202135}
\bibinfo{author}{RB, A.} \& \bibinfo{author}{KR, S.~K.}
\newblock \bibinfo{title}{Credit card fraud detection using artificial neural
  network}.
\newblock \emph{\bibinfo{journal}{Global Transitions Proceedings}}
  \textbf{\bibinfo{volume}{2}}, \bibinfo{pages}{35--41} (\bibinfo{year}{2021}).
\newblock
  \urlprefix\url{https://www.sciencedirect.com/science/article/pii/S2666285X21000066}.
\newblock \bibinfo{note}{1st International Conference on Advances in
  Information, Computing and Trends in Data Engineering (AICDE - 2020)}.

\bibitem{10.1007/978-3-030-89654-6_2}
\bibinfo{author}{Akande, O.~N.}, \bibinfo{author}{Misra, S.},
  \bibinfo{author}{Akande, H.~B.}, \bibinfo{author}{Oluranti, J.} \&
  \bibinfo{author}{Damasevicius, R.}
\newblock \bibinfo{editor}{Florez, H.} \& \bibinfo{editor}{Pollo-Cattaneo,
  M.~F.} (eds) \emph{\bibinfo{title}{A supervised approach to credit card fraud
  detection using an artificial neural network}}.
\newblock (eds \bibinfo{editor}{Florez, H.} \& \bibinfo{editor}{Pollo-Cattaneo,
  M.~F.}) \emph{\bibinfo{booktitle}{Applied Informatics}},
  \bibinfo{pages}{13--25} (\bibinfo{publisher}{Springer International
  Publishing}, \bibinfo{address}{Cham}, \bibinfo{year}{2021}).

\bibitem{Ileberi2022}
\bibinfo{author}{Ileberi, E.}, \bibinfo{author}{Sun, Y.} \&
  \bibinfo{author}{Wang, Z.}
\newblock \bibinfo{title}{A machine learning based credit card fraud detection
  using the ga algorithm for feature selection}.
\newblock \emph{\bibinfo{journal}{Journal of Big Data}}
  \textbf{\bibinfo{volume}{9}}, \bibinfo{pages}{24} (\bibinfo{year}{2022}).
\newblock \urlprefix\url{https://doi.org/10.1186/s40537-022-00573-8}.

\bibitem{10081315}
\bibinfo{author}{Mienye, I.~D.} \& \bibinfo{author}{Sun, Y.}
\newblock \bibinfo{title}{A deep learning ensemble with data resampling for
  credit card fraud detection}.
\newblock \emph{\bibinfo{journal}{IEEE Access}} \textbf{\bibinfo{volume}{11}},
  \bibinfo{pages}{30628--30638} (\bibinfo{year}{2023}).

\bibitem{10.1145/2907070}
\bibinfo{author}{Branco, P.}, \bibinfo{author}{Torgo, L.} \&
  \bibinfo{author}{Ribeiro, R.~P.}
\newblock \bibinfo{title}{A survey of predictive modeling on imbalanced
  domains}.
\newblock \emph{\bibinfo{journal}{ACM Comput. Surv.}}
  \textbf{\bibinfo{volume}{49}} (\bibinfo{year}{2016}).
\newblock \urlprefix\url{https://doi.org/10.1145/2907070}.

\bibitem{doi:10.1080/0952813X.2021.1907795}
\bibinfo{author}{Amit~Singh, R. K.~R.} \& \bibinfo{author}{Tiwari, A.}
\newblock \bibinfo{title}{Credit card fraud detection under extreme imbalanced
  data: A comparative study of data-level algorithms}.
\newblock \emph{\bibinfo{journal}{Journal of Experimental \& Theoretical
  Artificial Intelligence}} \textbf{\bibinfo{volume}{34}},
  \bibinfo{pages}{571--598} (\bibinfo{year}{2022}).
\newblock \urlprefix\url{https://doi.org/10.1080/0952813X.2021.1907795}.

\bibitem{9698195}
\bibinfo{author}{Esenogho, E.}, \bibinfo{author}{Mienye, I.~D.},
  \bibinfo{author}{Swart, T.~G.}, \bibinfo{author}{Aruleba, K.} \&
  \bibinfo{author}{Obaido, G.}
\newblock \bibinfo{title}{A neural network ensemble with feature engineering
  for improved credit card fraud detection}.
\newblock \emph{\bibinfo{journal}{IEEE Access}} \textbf{\bibinfo{volume}{10}},
  \bibinfo{pages}{16400--16407} (\bibinfo{year}{2022}).

\bibitem{Yang2022}
\bibinfo{author}{Yang, F.} \emph{et~al.}
\newblock \bibinfo{title}{A hybrid sampling algorithm combining synthetic
  minority over-sampling technique and edited nearest neighbor for missed
  abortion diagnosis}.
\newblock \emph{\bibinfo{journal}{BMC Medical Informatics and Decision Making}}
  \textbf{\bibinfo{volume}{22}}, \bibinfo{pages}{344} (\bibinfo{year}{2022}).
\newblock \urlprefix\url{https://doi.org/10.1186/s12911-022-02075-2}.

\bibitem{Zhang2023}
\bibinfo{author}{Zhang, Z.-L.}, \bibinfo{author}{Peng, R.-R.},
  \bibinfo{author}{Ruan, Y.-P.}, \bibinfo{author}{Wu, J.} \&
  \bibinfo{author}{Luo, X.-G.}
\newblock \bibinfo{title}{Esmote: an overproduce-and-choose synthetic examples
  generation strategy based on evolutionary computation}.
\newblock \emph{\bibinfo{journal}{Neural Computing and Applications}}
  \textbf{\bibinfo{volume}{35}}, \bibinfo{pages}{6891--6977}
  (\bibinfo{year}{2023}).
\newblock \urlprefix\url{https://doi.org/10.1007/s00521-022-08004-8}.

\bibitem{ebiaredoh2022machine}
\bibinfo{author}{Ebiaredoh-Mienye, S.~A.}, \bibinfo{author}{Swart, T.~G.},
  \bibinfo{author}{Esenogho, E.} \& \bibinfo{author}{Mienye, I.~D.}
\newblock \bibinfo{title}{A machine learning method with filter-based feature
  selection for improved prediction of chronic kidney disease}.
\newblock \emph{\bibinfo{journal}{Bioengineering}}
  \textbf{\bibinfo{volume}{9}}, \bibinfo{pages}{350} (\bibinfo{year}{2022}).

\bibitem{9204584}
\bibinfo{author}{Cheng, D.}, \bibinfo{author}{Wang, X.},
  \bibinfo{author}{Zhang, Y.} \& \bibinfo{author}{Zhang, L.}
\newblock \bibinfo{title}{Graph neural network for fraud detection via
  spatial-temporal attention}.
\newblock \emph{\bibinfo{journal}{IEEE Transactions on Knowledge and Data
  Engineering}} \textbf{\bibinfo{volume}{34}}, \bibinfo{pages}{3800--3813}
  (\bibinfo{year}{2022}).

\bibitem{sheng2023semi}
\bibinfo{author}{Sheng, X.}, \bibinfo{author}{Li, Y.}, \bibinfo{author}{Liu,
  Z.} \& \bibinfo{author}{Sun, M.}
\newblock \emph{\bibinfo{title}{Semi-supervised credit card fraud detection via
  attribute-driven graph representation}}, Vol.~\bibinfo{volume}{37},
  \bibinfo{pages}{1234--1241} (\bibinfo{year}{2023}).

\bibitem{9724422}
\bibinfo{author}{Ling, Y.}, \bibinfo{author}{Zhang, R.}, \bibinfo{author}{Cen,
  M.}, \bibinfo{author}{Wang, X.} \& \bibinfo{author}{Jiang, M.}
\newblock \emph{\bibinfo{title}{Cost-sensitive heterogeneous integration for
  credit card fraud detection}}, \bibinfo{pages}{750--757}
  (\bibinfo{year}{2021}).

\bibitem{zhang2023expressive}
\bibinfo{author}{Zhang, B.} \emph{et~al.}
\newblock \bibinfo{title}{The expressive power of graph neural networks: A
  survey} (\bibinfo{year}{2023}).
\newblock \eprint{2308.08235}.

\bibitem{hamilton2018inductive}
\bibinfo{author}{Hamilton, W.~L.}, \bibinfo{author}{Ying, R.} \&
  \bibinfo{author}{Leskovec, J.}
\newblock \bibinfo{title}{Inductive representation learning on large graphs}
  (\bibinfo{year}{2018}).
\newblock \eprint{1706.02216}.

\bibitem{vel2018graph}
\bibinfo{author}{Veličković, P.} \emph{et~al.}
\newblock \bibinfo{title}{Graph attention networks} (\bibinfo{year}{2018}).
\newblock \eprint{1710.10903}.

\bibitem{10.1145/3442381.3449989}
\bibinfo{author}{Liu, Y.} \emph{et~al.}
\newblock \emph{\bibinfo{title}{Pick and choose: A gnn-based imbalanced
  learning approach for fraud detection}}, WWW '21,
  \bibinfo{pages}{3168–3177} (\bibinfo{publisher}{Association for Computing
  Machinery}, \bibinfo{address}{New York, NY, USA}, \bibinfo{year}{2021}).
\newblock \urlprefix\url{https://doi.org/10.1145/3442381.3449989}.

\bibitem{10.1145/3397271.3401253}
\bibinfo{author}{Liu, Z.}, \bibinfo{author}{Dou, Y.}, \bibinfo{author}{Yu,
  P.~S.}, \bibinfo{author}{Deng, Y.} \& \bibinfo{author}{Peng, H.}
\newblock \emph{\bibinfo{title}{Alleviating the inconsistency problem of
  applying graph neural network to fraud detection}}, SIGIR '20,
  \bibinfo{pages}{1569–1572} (\bibinfo{publisher}{Association for Computing
  Machinery}, \bibinfo{address}{New York, NY, USA}, \bibinfo{year}{2020}).
\newblock \urlprefix\url{https://doi.org/10.1145/3397271.3401253}.

\bibitem{10.1145/3623401}
\bibinfo{author}{Tang, H.}, \bibinfo{author}{Wang, C.}, \bibinfo{author}{Zheng,
  J.} \& \bibinfo{author}{Jiang, C.}
\newblock \bibinfo{title}{Enabling graph neural networks for semi-supervised
  risk prediction in online credit loan services}.
\newblock \emph{\bibinfo{journal}{ACM Trans. Intell. Syst. Technol.}}
  (\bibinfo{year}{2023}).
\newblock \urlprefix\url{https://doi.org/10.1145/3623401}.
\newblock \bibinfo{note}{Just Accepted}.

\bibitem{10.1145/3269206.3272010}
\bibinfo{author}{Liu, Z.} \emph{et~al.}
\newblock \emph{\bibinfo{title}{Heterogeneous graph neural networks for
  malicious account detection}}, CIKM '18, \bibinfo{pages}{2077–2085}
  (\bibinfo{publisher}{Association for Computing Machinery},
  \bibinfo{address}{New York, NY, USA}, \bibinfo{year}{2018}).
\newblock \urlprefix\url{https://doi.org/10.1145/3269206.3272010}.

\bibitem{10.14778/3494124.3494128}
\bibinfo{author}{Rao, S.~X.} \emph{et~al.}
\newblock \bibinfo{title}{Xfraud: Explainable fraud transaction detection}.
\newblock \emph{\bibinfo{journal}{Proc. VLDB Endow.}}
  \textbf{\bibinfo{volume}{15}}, \bibinfo{pages}{427–436}
  (\bibinfo{year}{2021}).
\newblock \urlprefix\url{https://doi.org/10.14778/3494124.3494128}.

\bibitem{VANBELLE2022116463}
\bibinfo{author}{{Van Belle}, R.}, \bibinfo{author}{{Van Damme}, C.},
  \bibinfo{author}{Tytgat, H.} \& \bibinfo{author}{{De Weerdt}, J.}
\newblock \bibinfo{title}{Inductive graph representation learning for fraud
  detection}.
\newblock \emph{\bibinfo{journal}{Expert Systems with Applications}}
  \textbf{\bibinfo{volume}{193}}, \bibinfo{pages}{116463}
  (\bibinfo{year}{2022}).
\newblock
  \urlprefix\url{https://www.sciencedirect.com/science/article/pii/S0957417421017449}.

\bibitem{9163376}
\bibinfo{author}{Tingfei, H.}, \bibinfo{author}{Guangquan, C.} \&
  \bibinfo{author}{Kuihua, H.}
\newblock \bibinfo{title}{Using variational auto encoding in credit card fraud
  detection}.
\newblock \emph{\bibinfo{journal}{IEEE Access}} \textbf{\bibinfo{volume}{8}},
  \bibinfo{pages}{149841--149853} (\bibinfo{year}{2020}).

\bibitem{ebiaredoh2021artificial}
\bibinfo{author}{Ebiaredoh-Mienye, S.~A.}, \bibinfo{author}{Esenogho, E.} \&
  \bibinfo{author}{Swart, T.~G.}
\newblock \bibinfo{title}{Artificial neural network technique for improving
  prediction of credit card default: A stacked sparse autoencoder approach}.
\newblock \emph{\bibinfo{journal}{International Journal of Electrical and
  Computer Engineering}} \textbf{\bibinfo{volume}{11}}, \bibinfo{pages}{4392}
  (\bibinfo{year}{2021}).

\bibitem{ebiaredoh2020integrating}
\bibinfo{author}{Ebiaredoh-Mienye, S.~A.}, \bibinfo{author}{Esenogho, E.} \&
  \bibinfo{author}{Swart, T.~G.}
\newblock \bibinfo{title}{Integrating enhanced sparse autoencoder-based
  artificial neural network technique and softmax regression for medical
  diagnosis}.
\newblock \emph{\bibinfo{journal}{Electronics}} \textbf{\bibinfo{volume}{9}},
  \bibinfo{pages}{1963} (\bibinfo{year}{2020}).

\bibitem{10.1145/3366423.3380027}
\bibinfo{author}{Hu, Z.}, \bibinfo{author}{Dong, Y.}, \bibinfo{author}{Wang,
  K.} \& \bibinfo{author}{Sun, Y.}
\newblock \emph{\bibinfo{title}{Heterogeneous graph transformer}}, WWW '20,
  \bibinfo{pages}{2704–2710} (\bibinfo{publisher}{Association for Computing
  Machinery}, \bibinfo{address}{New York, NY, USA}, \bibinfo{year}{2020}).
\newblock \urlprefix\url{https://doi.org/10.1145/3366423.3380027}.

\bibitem{NIPS2017_3f5ee243}
\bibinfo{author}{Vaswani, A.} \emph{et~al.}
\newblock \bibinfo{editor}{Guyon, I.} \emph{et~al.} (eds)
  \emph{\bibinfo{title}{Attention is all you need}}.
\newblock (eds \bibinfo{editor}{Guyon, I.} \emph{et~al.})
  \emph{\bibinfo{booktitle}{Advances in Neural Information Processing
  Systems}}, Vol.~\bibinfo{volume}{30} (\bibinfo{publisher}{Curran Associates,
  Inc.}, \bibinfo{year}{2017}).
\newblock
  \urlprefix\url{https://proceedings.neurips.cc/paper_files/paper/2017/file/3f5ee243547dee91fbd053c1c4a845aa-Paper.pdf}.

\bibitem{Kingma_2019}
\bibinfo{author}{Kingma, D.~P.} \& \bibinfo{author}{Welling, M.}
\newblock \bibinfo{title}{An introduction to variational autoencoders}.
\newblock \emph{\bibinfo{journal}{Foundations and Trends{\textregistered} in
  Machine Learning}} \textbf{\bibinfo{volume}{12}}, \bibinfo{pages}{307--392}
  (\bibinfo{year}{2019}).
\newblock \urlprefix\url{https://doi.org/10.1561%2F2200000056}.

\bibitem{Powers2011EvaluationFP}
\bibinfo{author}{Powers, D. M.~W.}
\newblock \bibinfo{title}{Evaluation: from precision, recall and f-measure to
  roc, informedness, markedness and correlation}.
\newblock \emph{\bibinfo{journal}{ArXiv}}
  \textbf{\bibinfo{volume}{abs/2010.16061}} (\bibinfo{year}{2011}).
\newblock \urlprefix\url{https://api.semanticscholar.org/CorpusID:3770261}.

\bibitem{neptune2023}
\bibinfo{author}{Neptune.ai}.
\newblock \bibinfo{title}{F1 score, accuracy, roc auc, pr auc: Evaluation
  metrics you need to know}.
\newblock
  \bibinfo{howpublished}{\url{https://neptune.ai/blog/f1-score-accuracy-roc-auc-pr-auc}}
  (\bibinfo{year}{2023}).
\newblock \bibinfo{note}{Accessed: 2024-05-25}.

\bibitem{Kartik2021}
\bibinfo{author}{Kartik2112}.
\newblock \bibinfo{title}{Credit card transactions fraud detection dataset}.
\newblock \bibinfo{howpublished}{Retrieved 2023-09-21 from
  \url{https://www.kaggle.com/datasets/kartik2112/fraud-detection}}
  (\bibinfo{year}{2021}).

\bibitem{Arora2020}
\bibinfo{author}{Arora, V.}, \bibinfo{author}{Leekha, R.~S.},
  \bibinfo{author}{Lee, K.} \& \bibinfo{author}{Kataria, A.}
\newblock \bibinfo{title}{Facilitating user authorization from imbalanced data
  logs of credit cards using artificial intelligence}.
\newblock \emph{\bibinfo{journal}{Mobile Information Systems}}
  \textbf{\bibinfo{volume}{2020}}, \bibinfo{pages}{8885269}
  (\bibinfo{year}{2020}).
\newblock \urlprefix\url{https://doi.org/10.1155/2020/8885269}.

\end{thebibliography}

\end{document}